\documentclass[10pt,twocolumn,letterpaper]{article}

\usepackage{iccv}
\usepackage{times}
\usepackage{epsfig}
\usepackage{graphicx}
\usepackage{amsmath}
\usepackage{amssymb}
\usepackage{multirow}

\usepackage[pagebackref=true,breaklinks=true,colorlinks=true,bookmarks=false,linkcolor={red!60!black},urlcolor={magenta!80!black},citecolor={green!60!black}]{hyperref} 
\usepackage[noadjust]{cite}
\usepackage{arydshln}
\iccvfinalcopy %

\usepackage{xspace}
\usepackage{booktabs}
\RequirePackage{xspace}
\usepackage{upgreek}
\usepackage{color}
\usepackage[dvipsnames]{xcolor}
\usepackage[labelsep=period]{caption}
\usepackage{enumitem}
\usepackage{authblk}

\usepackage[final]{animate}
\usepackage{multirow}
\usepackage{subfig}

\def\eg{\emph{e.g}\onedot} 
\def\ie{\emph{i.e}\onedot} 
 
\def\etc{\emph{etc}\onedot} 
\def\wrt{w.r.t\onedot} 
\def\etal{\emph{et al}\onedot}

\newcommand{\Tref}[1]{Table~\ref{#1}}

\newcommand{\fref}[1]{Fig.~\ref{#1}}
\newcommand{\Fref}[1]{Figure~\ref{#1}}

\newcommand{\RadField}[0]{canonical radiance field\xspace}
\newcommand{\DefField}[0]{canonical trajectory field\xspace}

\newcommand{\InpaintNet}[0]{inpaint network\xspace}

\newcommand{\RadNet}{F_{\theta_1}}
\newcommand{\DefNet}{F_{\theta_2}}
\newcommand{\Warper}{F_{\text{warp}}}
\newcommand{\InNet}{F_{\theta_3}}
\newcommand{\Renderer}{F_{\text{render}}}
\newcommand{\ViewNet}{F_{\theta_4}}

\newcommand{\Vgrid}[2]{\mathbf{V}_{#1}^{#2}}
\newcommand{\decay}[1]{\alpha \left( #1 \right)}
\newcommand{\Loss}[1]{\mathcal{L}^{\text{#1}}}

\newcommand{\tabfirst}[1]{\textcolor{BrickRed}{\textbf{#1}}}
\newcommand{\tabsecond}[1]{\textcolor{BlueViolet}{\textbf{#1}}}

\renewcommand{\paragraph}[1]{\vspace{0.2em}\noindent \textbf{#1 \hspace{0.2em}}}

\def\iccvPaperID{10050} %

\ificcvfinal\fi

\title{Forward Flow for Novel View Synthesis of Dynamic Scenes}

\ificcvfinal
\usepackage{authblk}

\makeatletter
\renewcommand\AB@affilsepx{, \protect\Affilfont}
\makeatother
\author[1]{Xiang Guo}
\author[1,3]{Jiadai Sun}
\author[1]{Yuchao Dai$^{\dagger}$}
\author[2]{Guanying Chen}
\author[3]{Xiaoqing Ye}
\author[3]{Xiao Tan}
\author[3]{Errui Ding}
\author[3]{Yumeng Zhang}
\author[3]{Jingdong Wang}
\affil[1]{Northwestern Polytechnical University}
\affil[2]{FNii and SSE, CUHK-Shenzhen}
\affil[3]{Baidu Inc.}
\fi

\ificcvfinal\fi

\usepackage[accsupp]{axessibility}  %

\begin{document}

\maketitle

\ificcvfinal\thispagestyle{empty}\fi

\noindent\let\thefootnote\relax\footnotetext{{${\dagger}$ Corresponding author \tt(daiyuchao@nwpu.edu.cn).}
The first three authors also with the Shaanxi Key Laboratory of Information Acquisition and Processing.
}

\begin{abstract}
This paper proposes a neural radiance field (NeRF) approach for novel view synthesis of dynamic scenes using forward warping.
Existing methods often adopt a static NeRF to represent the canonical space, and render dynamic images at other time steps by mapping the sampled 3D points back to the canonical space with the learned \emph{backward flow} field. 
However, this backward flow field is non-smooth and discontinuous, which is difficult to be fitted by commonly used smooth motion models. 
To address this problem, we propose to estimate the \emph{forward flow} field and directly warp the canonical radiance field to other time steps. Such forward flow field is smooth and continuous within the object region, which benefits the motion model learning. 
To achieve this goal, we represent the canonical radiance field with voxel grids to enable efficient forward warping, and propose a differentiable warping process, including an average splatting operation and an inpaint network, to resolve the many-to-one and one-to-many mapping issues.
Thorough experiments show that our method outperforms existing methods in both novel view rendering and motion modeling, demonstrating the effectiveness of our forward flow motion modeling. Project page:  \url{https://npucvr.github.io/ForwardFlowDNeRF}.

\end{abstract}

\section{Introduction}\label{sec:introduction}

Novel view synthesis (NVS) is a challenging and long-standing problem in computer vision and graphics, which has many applications in virtual reality, augmented reality, data augmentation, image editing, \etc. Recently, differentiable neural rendering~\cite{mildenhall2020_nerf_eccv20,yariv2020multiview,niemeyer2020differentiable} has been introduced into this area. 
In particular, the neural radiance field (NeRF) \cite{mildenhall2020_nerf_eccv20} promotes this area significantly and attracts lots of interest within a short time. NeRF~\cite{mildenhall2020_nerf_eccv20} produces realistic images by representing the 3D world with a multi-layer perceptron (MLP), which maps the input 3D coordinates and 2D view direction to target density and color.

\begin{figure}[t] \centering
    \includegraphics[width=\columnwidth]{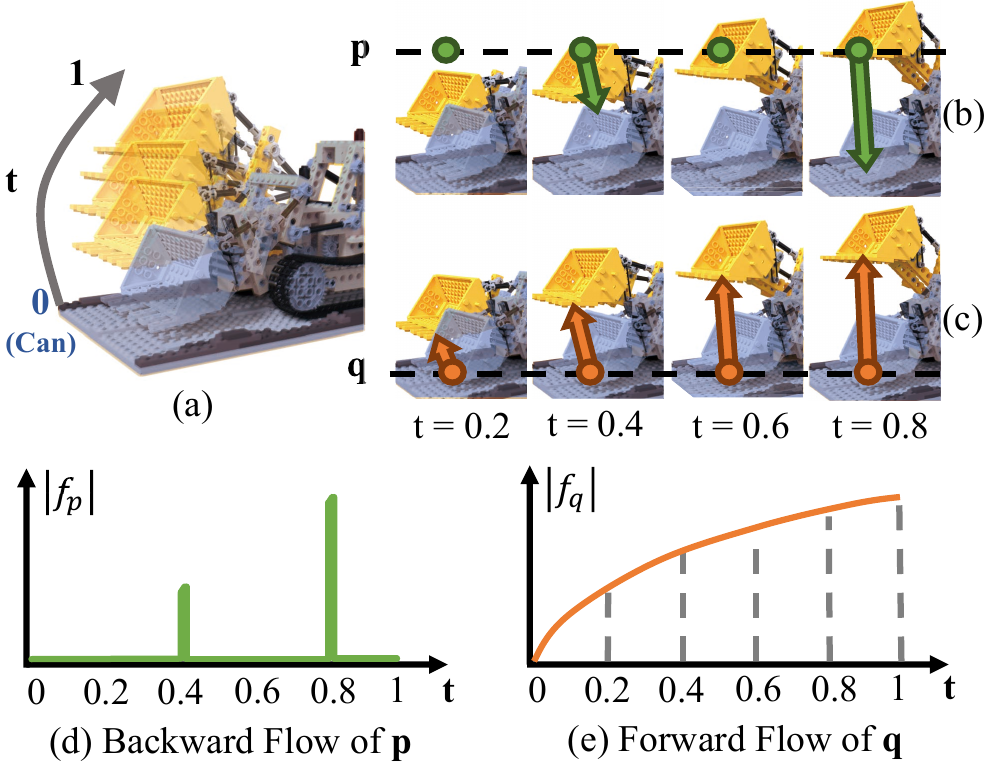}
    \caption{\textbf{Comparison of backward flow and forward flow.}
    This figure shows an example of backward and forward flow changes. \textbf{(a)} An example of dynamic scene. \textbf{(b)} With the bucket lifting up, different types of points cover the green point $\mathbf{p}$, which needs very different backward flows to map this point back to canonical space. \textbf{(d)} shows the norm changes of the backward flow, which is not smooth. \textbf{(c)} On the other hand, the forward flow of position $\mathbf{q}$, which maps the constant object point from canonical space to other times, is smooth and continuous. \textbf{(e)} shows the norm changes of the forward flow.
    } \label{fig:backvsforward}
    \vspace{-2\baselineskip}
\end{figure}

While the original NeRF~\cite{mildenhall2020_nerf_eccv20} can only model static scenes, a series of works extend the NeRF-based framework from static to dynamic scenes \cite{gao2021dynamic_iccv21,li2021_nsff_cvpr21,xian2021_space_cvpr21,tretschk2020_nonrigid_iccv21,park2021_nerfies_iccv21,pumarola2021_dnerf_cvpr21,wang2021_dctnerf_arxiv,du2021_nerflow_iccv21}. 
One of the promising directions is using a canonical space representation~\cite{tretschk2020_nonrigid_iccv21, pumarola2021_dnerf_cvpr21, guo2022_NDVG_arxiv}. 
This representation sets one of the time steps as canonical time and models the static scene with a canonical radiance field. 
To render images at other time steps, a deformation field is used to estimate the \emph{backward flow} for moving the 3D points from the current time step back to the canonical time step. 
Although the canonical-based representation with the backward flow is easy to implement, the backward flow field is non-smooth and discontinuous. 
As shown in \fref{fig:backvsforward}(b), for a fixed 3D position along the timeline, there will be different types of points covering the position $\mathbf{p}$, which needs discontinuous flows to map them back to canonical space (\fref{fig:backvsforward}(d)). So the backward flow cannot be fitted well with commonly used smooth motion models (MLP, for example). Also, distortions are introduced to the static canonical space geometry due to the failure of the motion model, as shown in \fref{fig:cancompare}.

To solve the problem of backward flow, we propose using \emph{forward flow} as the deformation model. Instead of warping the sampled points on image rays at other time steps back to the canonical time and rendering at the canonical space, we propose to warp the whole canonical radiance field from the canonical time to other time steps using forward deformation flow and render at corresponding time steps. In this way, the forward flow estimated by the deformation model will be smooth and continuous for the same 3D position along the timeline (\fref{fig:backvsforward}(c) and (e)). 
Note that SNARF~\cite{chen2021_snarf_ICCV} has also used forward warping based on an inverse skinning model, but it is designed for dynamic human modeling and cannot be used in general scenes. In this paper, we aim to achieve forward warping for general scenes, which means we must warp the whole space.

However, introducing forward warping into the canonical space based NeRF methods is not straightforward as there are three main problems to be solved.
First, the traditional canonical radiance field in existing methods cannot be warped explicitly, since the radiance field is represented as a continuous function parameterized by an MLP. 
To solve this problem, we propose to use the voxel grid to represent the canonical radiance field as it is finite and discrete.
Recent voxel-based methods~\cite{yu2021_plenoctrees_iccv21,sun2021direct,muller2022instant} have proven the effectiveness of this representation. 
The other two problems are the \emph{many-to-one} and \emph{one-to-many} mapping issues brought by the inherent property of the forward warping operation.
To address them, we propose a \emph{differentiable forward warping} method consisting of an \emph{average splatting} operation and an \emph{inpaint network} to solve the \emph{many-to-one} and \emph{one-to-many} issue, respectively. Extensive experiments have been conducted to verify the effectiveness of our method.

Our key contributions can be summarized as follows:
\begin{itemize}[itemsep=0pt,parsep=0pt,topsep=2bp]
    \item To the best of our knowledge, we are the first to investigate forward warping in dynamic view synthesis for general scenes. We propose a novel canonical based NeRF with forward flow motion modeling for dynamic view synthesis. Thanks to the forward flow field, our method can better represent the object motions, and explicitly recover the trajectory of a surface point.
    \item We introduce voxel grid based canonical radiance field to enable reasonable computation of forward warping, and propose a differentiable forward warping method, including an average splatting operation and an inpaint network, to solve the many-to-one and one-to-many issues of forward warping.
    \item Experiments on multiple datasets show that our method outperforms existing methods on the D-NeRF~\cite{pumarola2021_dnerf_cvpr21} dataset, Hypernerf~\cite{park2021hypernerf} dataset, NHR~\cite{wu2020_NHR} dataset and our proposed dataset.
\end{itemize}

\section{Related Works}\label{sec:relatedwork}

\begin{figure*}[t] \centering
    \includegraphics[width=0.95\textwidth]{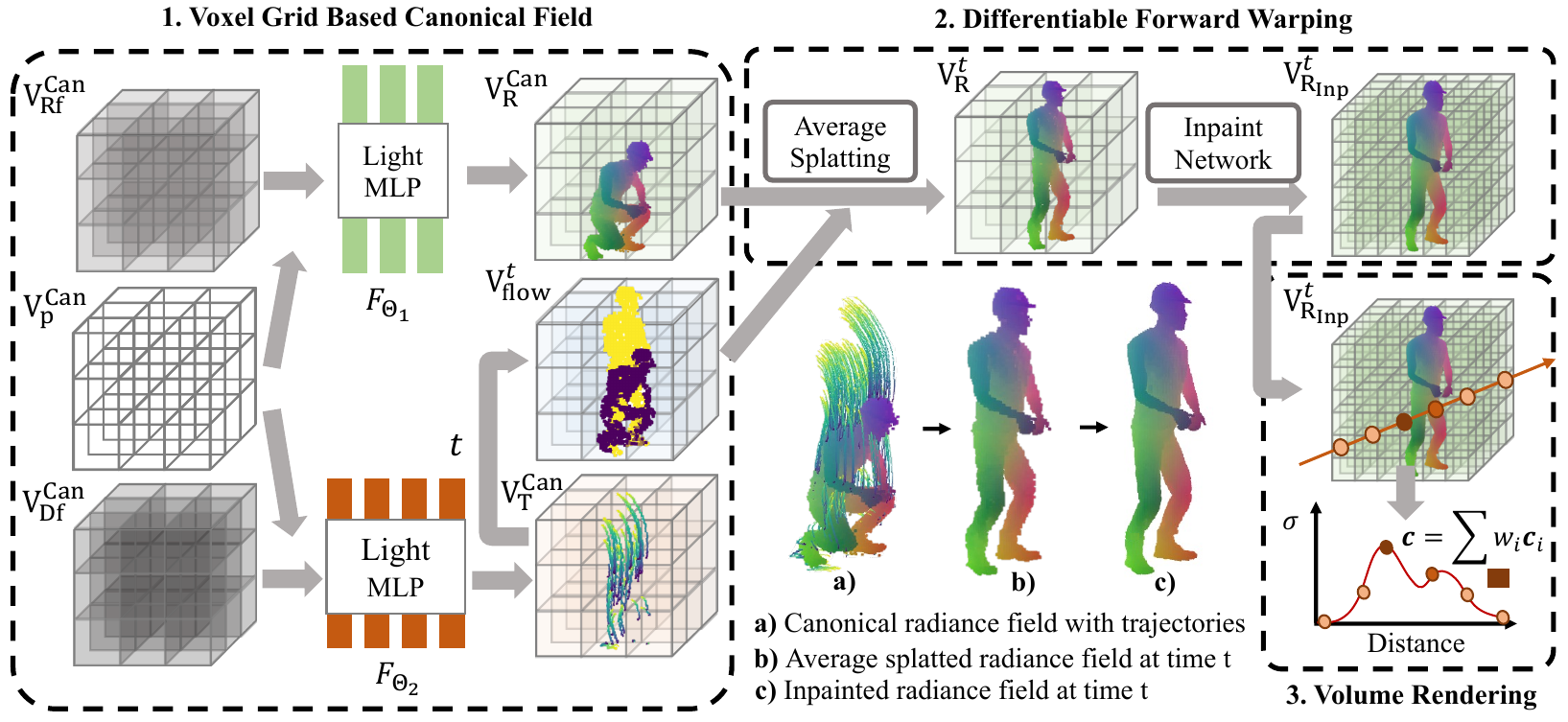}
    \caption{\textbf{Overview of our proposed method.} \textbf{a)} We represent a static scene at canonical time with a voxel grid based radiance field for density\&color and a voxel grid based trajectory field for deformations;  \textbf{b)} We propose to first forward warp canonical radiance field using the forward flow by average splatting; \textbf{c)} We then inpaint the warped radiance field using a inpaint network; Specifically, \textbf{1. Voxel Grid Based Canonical Field} contains two models. The canonical radiance field $\Vgrid{\text{R}}{\text{Can}}$ is estimated by a Light MLP which takes canonical radiance feature $\Vgrid{\text{Rf}}{\text{Can}}$ and corresponding 3D coordinates $\Vgrid{\text{p}}{\text{Can}}$ as input. The canonical trajectory field $\Vgrid{\text{T}}{\text{Can}}$ is estimated by another Light MLP which takes deformation feature and coordinates as input. The deformation flow $\Vgrid{\text{flow}}{t}$ from canonical to time $t$ can then be obtained; \textbf{2. Differential Forward Warping} first warp $\Vgrid{\text{R}}{\text{Can}}$ to get radiance field $\Vgrid{\text{R}}{t}$ at time $t$. Then, the $\Vgrid{\text{R}}{t}$ is inpainted by a inpaint network, which is $\Vgrid{\text{R}_{\text{Inp}}}{t}$; \textbf{3. Volume Rendering} render colors of rays at time $t$ based on $\Vgrid{\text{R}_{\text{Inp}}}{t}$} 
    \label{fig:pipeline}
    \vspace{-2\baselineskip}
\end{figure*}

Novel View Synthesis (NVS) is a long-standing task in both computer vision and graphics~\cite{buehler2001_unstructured,chen1993view,levoy1996light,greene1986environment}, and surveys of recent methods can be found in~\cite{shum2000review,tewari2020state,tewari2021advances}.
Some methods require to reconstruct an explicit 3D model to represent the scene, such as point clouds~\cite{aliev2020_neural_eccv20}, voxels or meshes~\cite{riegler2020_freenvs_eccv20,riegler2021_stablenvs_cvpr21,thies2019_deferred_tog19,hedman2018_deepibr_tog18}. 
Then novel images from arbitrary views can be rendered from this geometry.
Another methods try to estimate depth map through multi-view geometry, and then aggregate features from multiple frames through the co-visibility, such as~\cite{kalantari2016learning_tog16,penner2017_soft_tog17,choi2019_extreme_nvs_iccv19,riegler2020_freenvs_eccv20,riegler2021_stablenvs_cvpr21,flynn2016_deepstereo_cvpr16,xu2019_deep_tog19}.
Recently, neural implicit representations have shown great promise for novel view synthesis and 3D modeling.
In this section, we focus more on the neural implicit rendering methods and mainly summarize these schemes according to whether they can handle dynamic scenes.

\paragraph{NVS for Static Scenes} NeRF~\cite{mildenhall2020_nerf_eccv20}, as a seminal work, uses MLPs to model a 5D radiance field, which can render impressive view synthesis for static scenes captured. 
Numerous subsequent works have extended NeRF to kinds of scenarios, such as larger and unbounded scenes~\cite{Zhang20arxiv_nerf++,tancik2022_blocknerf_cvpr,xiangli2022bungeenerf_eccv,Rematas2022_urbanNerf_CVPR,martin2021_nerfw_cvpr21},
relighting~\cite{boss2021nerd,srinivasan2021nerv,zhang2021nerfactor,yang2022s}, incorporating anti-aliasing for multi-scale rendering~\cite{barron2021mipnerf}, and generalization ability~\cite{chen2021mvsnerf,trevithick2021grf,yu2021_pixelnerf_cvpr21,wang2021_ibrnet_cvpr21}.
In addition, some methods are devoted to more efficient neural rendering and optimization in NeRF-like framework, such as~\cite{neff2021_donerf_egsr21,lindell2021_autoint_cvpr21, piala2021terminerf, liu2020_nsvf_nips20,yu2021_plenoctrees_iccv21,lombardi2021mixture} focus more on efficient sampling along each ray for color accumulation,~\cite{rebain2021_derf_cvpr21, Reiser2021_kiloNeRF_iccv21} subdivide the scene into multiple cells for efficient processing, and~\cite{yu2021plenoxels,sun2021direct,muller2022instant} exploit voxel-grid representation to speed up the optimization of radiance field.
However, these methods are mainly applicable to static scenes, leaving out the scenes with dynamic objects, which are actually more extensive and practical.

\paragraph{NVS for Dynamic Scenes} 
There are several works that extend NeRF from static scenes to dynamic scenes with non-rigid deformable objects.
One feasible way is to build a 4D spatial-temporal representation. 
For example, Yoon \etal \cite{yoon2020_nvidiadataset_cvpr20} combine single-view depth and depth from multi-view stereo to render virtual views with 3D warping. Gao \etal \cite{gao2021dynamic_iccv21} use a time-invariant model (static) and a time-varying model (dynamic) to represent the scenes, and regularize the dynamic model by scene flow estimation. NeRFlow~\cite{du2021_nerflow_iccv21} learns a 4D spatial-temporal representation of a dynamic scene from a set of RGB images. Xian \etal \cite{xian2021_space_cvpr21} build a 4D space-time irradiance field to map a spatial-temporal location to the emitted color and volume density. Similarly, NSFF~\cite{li2021_nsff_cvpr21} models the dynamic scene as a time-variant continuous function of appearance, geometry, and 3D scene motion.
DCT-NeRF~\cite{wang2021_dctnerf_arxiv} uses the Discrete Cosine Transform (DCT) to capture dynamic motion, \ie learning smooth and stable trajectories over time for each point in space. 

On the other hand, D-NeRF~\cite{pumarola2021_dnerf_cvpr21}, Nerfies \cite{park2021_nerfies_iccv21}, HyperNeRF~\cite{park2021hypernerf} and NR-NeRF~\cite{tretschk2020_nonrigid_iccv21} use a static canonical radiance field to capture geometry and appearance, and then learn a deformation/displacement field at each time step \wrt the canonical space.
Specifically, to render an image at an arbitrary time step, a deformation field is used to estimate backward scene flow, moving 3D points from the current time step back to the canonical step. However, for the same 3D location along the timeline, the backward flow field is not guaranteed to be smooth and continuous. 
As a result, the canonical geometry usually has distortions and resembles the mean shape of a moving object.
We focus on solving the problem of backward flow in this paper.

Along with the two main directions, there is a trend to speed up the training of dynamic NeRF, which is based on voxel grid representation.
TiNeuVox~\cite{fang2022_TANV_arxiv} models the deformation using a tiny MLP and uses multi-distance interpolation to get the feature for the radiance network which estimates the density and color. V4D~\cite{gan2022_V4D_arxiv} uses the 3D feature voxel to model the 4D radiance field with additional time dimension concatenated and proposes look-up tables for pixel-level refinement. Although V4D mainly focuses on improving image quality, 
the training speed is not significant compared to TiNeuVox.
DeVRF~\cite{liu2022_DeVRF_arxiv} also builds on voxel-grid representation, which proposes to use multi-view data to overcome the nontrivial problem of the monocular setup. Multi-view data simplifies the learning of motion and geometry compared with others using monocular images.

\section{Method}\label{sec:method}
\paragraph{Motivation} Backward-warping based methods~\cite{pumarola2021_dnerf_cvpr21, guo2022_NDVG_arxiv} propose a network $f_{t\rightarrow\text{Can}} = F(p, t)$ to estimate the deformation flow $f_{t\rightarrow\text{Can}}$ which moves the point at the position $p$ from other time steps $t$ back to canonical time $\operatorname{Can}$. However, for the same position $p$, at different time steps $t$s, there could be different object points or even the empty point at this position, as shown in \fref{fig:backvsforward}. This means the deformation flow $f_{t\rightarrow\text{Can}}$ is non-smooth and discontinuous along the timeline for a fixed position $p$. This could introduce difficulties for motion learning and produce distortion in the canonical radiance field, because the backward flow network has limited capacities in learning a correct non-smooth and discontinuous deformation field. On the other hand, our proposed forward warping strategy estimates the deformation $f_{\text{Can}\rightarrow t}$ from the canonical time to another time step of a 3D point in canonical space. The set of the deformation flows of one position along the timeline is actually the trajectory of this point. This guarantees the flows to be smooth and continuous along the timeline as we assume the motions in reality have these properties.

\paragraph{Method Overview}
We model a static scene at the canonical time using the voxel grid representation with a canonical radiance field and a canonical trajectory field. To synthesize dynamic images, we propose to forward warp the canonical radiance field to corresponding time steps and render the images using volume rendering based on the warped radiance field. \fref{fig:pipeline} shows the overview of our method. In the following, we will introduce three main components of our method: voxel grid based canonical field, differentiable forward warping, and volume rendering. Finally, the model optimization is represented, including the proposed loss functions and training strategy.

\subsection{Voxel Grid Based Canonical Field}

\paragraph{Canonical Radiance Field}
To warp a static scene in canonical time to other time steps using forward warping, the radiance field should be represented by finite 3D points. The original heavy MLP in NeRF \cite{mildenhall2020_nerf_eccv20} is not practical, as we cannot query infinite 3D points in the canonical space. Inspired by recent works \cite{yu2021plenoxels,sun2021direct,muller2022instant} on voxel grid presentations, we propose to use a learnable voxel radiance feature grid $\Vgrid{\text{Rf}}{\text{Can}}$ and a lightweight MLP network $\RadNet$ to model the radiance of the static scene as shown in \fref{fig:pipeline}. The \RadField is defined as
\begin{equation}
\Vgrid{\text{R}}{\text{Can}} = \RadNet(\Vgrid{\text{p}}{\text{Can}}, \Vgrid{\text{Rf}}{\text{Can}}),
\end{equation}
where $\Vgrid{\text{R}}{\text{Can}}= \{ \Vgrid{\sigma}{\text{Can}}, \Vgrid{\text{cf}}{\text{Can}} \}$ consists of density voxel grid $\Vgrid{\sigma}{\text{Can}}$ and color feature voxel grid $\Vgrid{\text{cf}}{\text{Can}}$. $\Vgrid{\text{p}}{\text{Can}}$ are the 3D coordinates of the voxel grid in canonical space, embedded according to NeRF \cite{mildenhall2020_nerf_eccv20}.

\begin{figure}[t] \centering
    \includegraphics[width=\columnwidth]{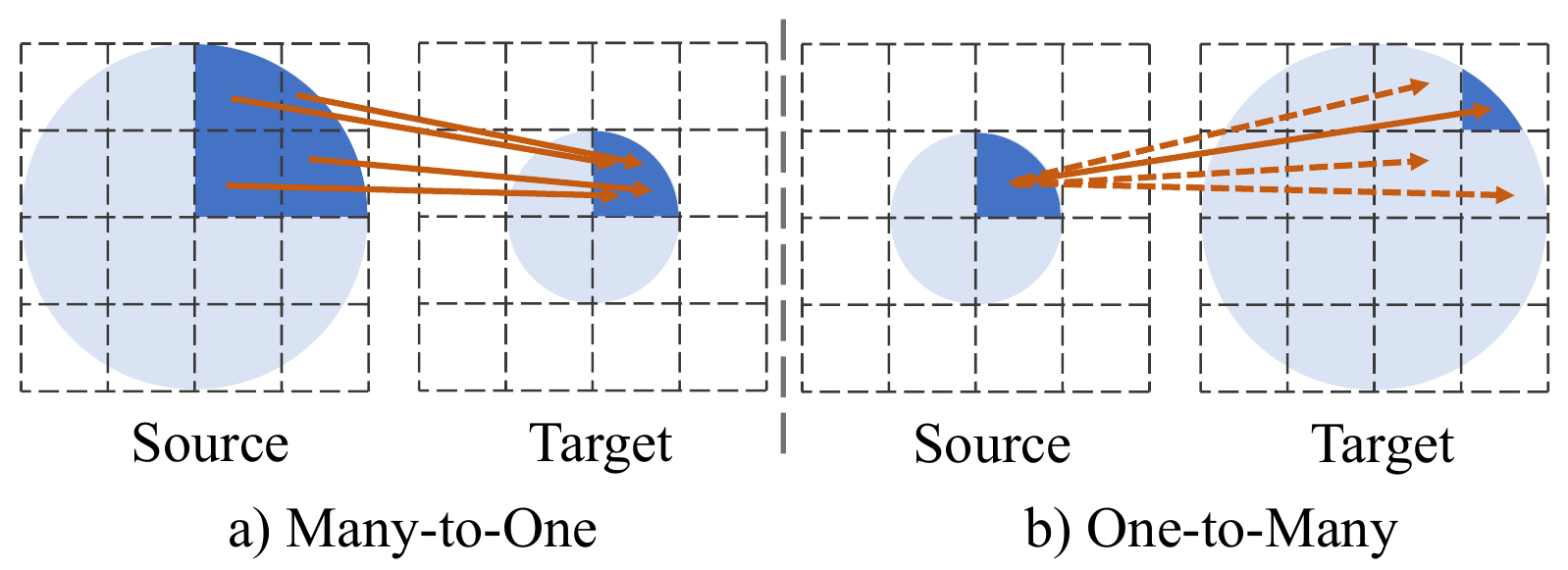}
    \caption{\textbf{Two issues of the forward warping.} Many-to-one: %
    multiple positions in the source correspond to the same position in the target. One-to-many: one position in the source corresponds to multiple positions in the target, causing holes if warping naively.} \label{fig:twoproblems}
    \vspace{-2\baselineskip}
\end{figure}

\paragraph{Canonical Trajectory Field}
We propose to use Discrete Cosine Transform (DCT) \cite{wang2021_dctnerf_arxiv} to represent the trajectory of a 3D point, which ensures the smoothness of the motion. Similar to \RadField, we also use a deformation feature grid $\Vgrid{\text{Df}}{\text{Can}}$ and a lightweight MLP network $\DefNet$ to model the \DefField, which is defined as
\begin{equation}
\Vgrid{T}{\text{Can}} = \DefNet(\Vgrid{\text{p}}{\text{Can}}, \Vgrid{\text{Df}}{\text{Can}}),
\end{equation}
where $\Vgrid{T}{\text{Can}}$ contains the DCT coefficients of the voxels. Given time step $t$, we could get the deformation flow of these voxels from canonical to time step $t$ by
\begin{equation}
\Vgrid{\text{flow}}{t} = f_{{\text{DCT}^{-1}}}(\Vgrid{T}{\text{Can}}, t) -  f_{{\text{DCT}^{-1}}}(\Vgrid{T}{\text{Can}}, \text{Can}),
\end{equation}
where $f_{{\text{DCT}^{-1}}}$ is the inverse DCT transform of the DCT coefficients as described in TrajectoryNeRF~ \cite{wang2021_dctnerf_arxiv}.

\subsection{Differentiable Forward Warping}
Forward warping has two major issues: many-to-one and one-to-many. \fref{fig:twoproblems} shows a simple example, where we aim to warp the source to the target using forward warping: a) Many-to-one: if a circle shrinks, there will be multiple positions corresponding to one position from the source to the target. b) One-to-many: if a circle expands, multiple positions in the target will correspond to the same position in the source, which leaves holes of the target if we warp the source to target naively. For the many-to-one issue, we propose to use the average splatting to fuse multiple values into one. For the one-to-many issue, we use an inpaint network to inpaint the missing positions of the warped grids. More details are in the following sections.

\paragraph{Average Splatting}
We propose to fuse possible multiple values from the source grid that are mapped into the same voxel position of the target grid, motivated by Softmax-splatting~\cite{Niklaus2020_VFI_CVPR}. Specifically, we propose a simple yet effective method that calculates the ``average" of these values with a trilinear kernel. Formally, suppose that we need to warp the source grid $\Vgrid{}{\text{S}}$ to target grid $\Vgrid{}{\text{T}}$ by the flow $f_{\text{S}\rightarrow\text{T}}$, and $\mathbf{p}$, $\mathbf{q}$ are indexes of a voxel grid.
We define $\Vgrid{}{\text{T}} = \Warper(\Vgrid{}{\text{S}}, f_{\text{S}\rightarrow\text{T}})$ as follows,
\begin{gather}
\Vgrid{}{\text{T}}[\mathbf{p}] = \frac{\sum_{\forall\mathbf{q}\in\Vgrid{}{\text{S}}}b[\mathbf{u}]\cdot\Vgrid{}{\text{S}}[\mathbf{q}]}{\sum_{\forall\mathbf{q}\in\Vgrid{}{\text{S}}}^{}b[\mathbf{u}]}, \\
b[\mathbf{u}]\!=\! \prod\text{max}(0, 1\!-\!\left|\mathbf{u}_i \right|), {i} \in \{x,y,z\}, \\
\mathbf{u} = (\mathbf{q}+f_{\text{S}\rightarrow\text{T}}[\mathbf{q}]) - \mathbf{p},
\end{gather}
where $x,y,z$ are three axes of the voxel grid, and $\mathbf{u}_i\in\mathbb{R}$ is one element of the vector $\mathbf{u}\in\mathbb{R}^3$.

\noindent So, we warp the canonical radiance field to time step $t$ by
\begin{equation}
\Vgrid{\text{R}}{t} = \Warper(\Vgrid{\text{R}}{\text{Can}},\Vgrid{\text{flow}}{t}).
\end{equation}

\begin{figure}[t] \centering
    \includegraphics[width=\columnwidth]{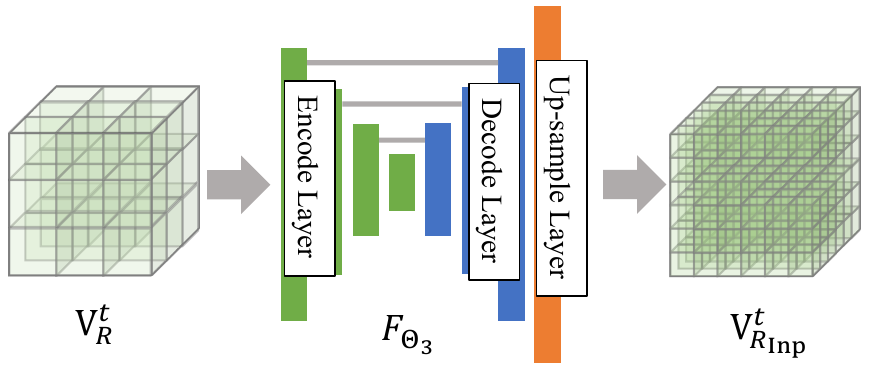}
    \caption{\textbf{Inpaint network.} It consists of a 3D U-Net structure and an up-sampling layer, which could inpaint and upsample the input voxel grid.} \label{fig:inpaintnet}
    \vspace{-2\baselineskip}
\end{figure}

\paragraph{Inpaint Network}
Another issue of forward warping is one-to-many, which leaves holes in the voxel grid after warping. To resolve this issue, we propose building an inpaint network to fill these holes. As shown in~\fref{fig:inpaintnet}, we modify a 3D U-Net structure network \cite{hardtke2020_3DUNet} followed by an up-sampling layer, which is $\Vgrid{\text{R}_{\text{Inp}}}{t}=\InNet(\Vgrid{\text{R}}{t})$, where $\Vgrid{\text{R}_{\text{Inp}}}{t}$ is the inpainted and up-sampled voxel grid.

This structure helps encoding layers to learn information from local neighboring voxels and decoding layers to recover the original resolution with filled content.
We only feed the warped voxel grid to the inpaint network, which has no access to temporal information. Thus, the inpaint network can not learn to warp, limiting its ability only to inpainting. Since warping operations and the inpaint network are relatively expensive to compute, we propose to attach an up-sample layer after the U-Net structure. In this way, we could warp the grids with a low resolution to save time and memory, while rendering with a higher resolution for better image quality. The up-sample layer consists of a interpolation layer, convolution layers, and activation layers.

\subsection{Volume Rendering}
After we get the radiance voxel grid at time $t$, the pixel colors of the image rays can be rendered using volume rendering techniques \cite{mildenhall2020_nerf_eccv20}. Given a ray $\mathbf{r}(w) = \mathbf{o} + w \mathbf{d}$ emitted from the camera center $\mathbf{o}$ with view direction $\mathbf{d}$ through a given pixel on the image plane, we render the corresponding pixel color $\mathbf{C_{\text{Inp}}}(\mathbf{r}) = \Renderer(\Vgrid{\text{R}_{\text{Inp}}}{t}, \mathbf{r})$. To do this, we get all 3D points $\mathbf{p}$ that the ray intersects with voxel grids. Next, 
trilinear interpolation is applied to obtain the density and color feature of each 3D point.
\begin{equation}
    (\sigma, \mathbf{c_f}) = F_{\text{inter}}(\Vgrid{\text{R}_{\text{Inp}}}{t}, \mathbf{p}).
\end{equation}
Then, we get the color of this 3D point $\mathbf{p}$ by
\begin{align}
    \mathbf{c} &=
        \begin{cases}
            \mathbf{c_f} \quad\quad\quad\quad \text{if $\mathbf{c_f} \in \mathbb{R}^3$},\\
            \ViewNet(\mathbf{c_f}, \mathbf{d}) \quad \text{otherwise},
        \end{cases}
\end{align}
where $\ViewNet$ is a small MLP network that is used to produce colors that depend on the direction of the ray. %

Finally, the pixel color could be rendered by
\begin{gather} %
    \mathbf{C}(\mathbf{r}) =  
        \sum\nolimits_{k=1}^{K} T(w_k) \, \decay{\sigma(w_k) \delta_{k}} \, \mathbf{c}(w_k) \, , \label{eqn:nerf1}\\
    T(w_k) = \exp \left(-\sum\nolimits_{j=1}^{k-1} \sigma(w_{j}) \delta_{j} \right) \, ,\label{eqn:nerf2}
\end{gather}
where $K$ is the number of sampled points along the ray, $\delta_{k}$ is the distance between adjacent samples along the ray, and $\decay{\sigma(w_k) \delta_{k}} = 1 - \exp (-\sigma(w_k) \delta_{k})$.

Note that the warped radiance grid $\Vgrid{\text{R}}{t}$ should also render reasonable images. So we directly up-sample $\Vgrid{\text{R}}{t}$ to get $\Vgrid{\text{R}_\text{Up}}{t}$ without inpainting, and render the corresponding pixel color $\mathbf{C_{\text{Up}}}(\mathbf{r}) \!=\! \Renderer(\Vgrid{\text{R}_{\text{Up}}}{t}, \mathbf{r})$.

\subsection{Model Optimization}
\noindent To optimize the model, we design a series of loss functions. 
\paragraph{Photometric Loss} First, the most important loss is the photometric loss, which is the mean square error (MSE) of the rendered color and ground truth color $\mathbf{C}_{\text{gt}}(\mathbf{r})$.
\begin{gather}
\begin{split}
    \Loss{photo} =& \frac{1}{|\mathcal{R}|} \sum_{r\in\mathcal{R}} \left\| \mathbf{C_{\text{Inp}}}(\mathbf{r}) - \mathbf{C}_{\text{gt}}(\mathbf{r}) \right\|_2^2 \\ 
    & + \frac{1}{|\mathcal{R}|} \sum_{r\in\mathcal{R}} \left\| \mathbf{C_{\text{Up}}}(\mathbf{r}) - \mathbf{C}_{\text{gt}}(\mathbf{r}) \right\|_2^2,
\end{split}
\end{gather}
where $\mathcal{R}$ is the set of rays sampled in one batch.

Following DVGO~\cite{sun2021direct}, we use $\Loss{ptc}$ to directly supervise the color of sampled points. The intuition is that sampled points with larger weights contribute more to the rendered color. Also, we use the background entropy loss $\Loss{bg}$ to encourage the densities concentrating on either the foreground or the background. 

\paragraph{Inpaint Network Loss} As the inpaint network is used to complete the warped voxel grids, the output of the inpaint network should be close to the input. Therefore, we propose to use the L1 norm of the difference between the input and output of the \InpaintNet.
\begin{gather}
\Loss{vdiff} =\left\| \Vgrid{\text{R}_{\text{Inp}}}{t} - \Vgrid{\text{R}_{\text{Up}}}{t} \right\|_1.
\end{gather}

\paragraph{Regularization Terms} Since modeling the appearance and motion of a dynamic scene from monocular images is a non-trivial problem, we propose a series of regularization terms.

First, we encourage most 3D points in canonical space to be static by introducing $\Loss{flow}$ to be the L1 norm of $\Vgrid{\text{flow}}{t}$.
Second, denoting $\Loss{tv}$ as the total variation function. We compute $\Loss{tv}(\Vgrid{\sigma}{\text{Can}})$ to ensure the spatial smoothness of the density in canonical space, and introduce $\Loss{tv}(\Vgrid{\text{flow}}{t})$ to encourage the spatial smoothness of the motions. 
Finally, inspired by RegNeRF~\cite{Niemeyer2022_RegNeRF_CVPR}, we render depths $\mathbf{D}$ of image patches sampled from random views and minimize their total variations $\Loss{tv}(\mathbf{D})$.

\paragraph{Overall Loss} The overall loss function used for optimization defined as follow, where $w_{1\sim7}$ are weights to balance each component in the final coarse loss.
\begin{gather}
\begin{split}
    \Loss{} =& \Loss{photo} + w_1\Loss{ptc} + w_2\Loss{bg} + w_3\Loss{flow} + w_4\Loss{vdiff}\\
    & + w_5\Loss{tv}(\Vgrid{\sigma}{\text{Can}}) + w_6\Loss{tv}(\Vgrid{\text{flow}}{t}) + w_7\Loss{tv}(\mathbf{D}),
\end{split}
\end{gather}

\paragraph{Training Strategy} %
First, we propose \emph{progressive training}, which starts training with images close to the canonical time and progressively adds images with farther time steps until all images are added. 
Second, we set up a \emph{coarse-to-fine training strategy}. In the coarse stage, we set the color feature to be the color itself, without considering the ray directions. We compute a smaller bounding box with the proxy geometry learned from the coarse stage, which could filter a large portion of empty space for the fine stage training. The model trained in the fine stage is our final model, and we set the color features to be high dimensional features and model the ray direction dependency with $\ViewNet$.
\section{Experiments} \label{sec:experiment}

\begin{table}[t!]
\setlength{\tabcolsep}{4pt} %
\centering
\caption{\small{
\textbf{Quantitative comparison on D-NeRF Dataset.} Comparison of our method with others on LPIPS and PSNR/SSIM. 
The \tabfirst{Red} text indicates the best and \tabsecond{blue} text is the second best result.
}}
\label{table:quant}
\footnotesize{%
\begin{tabular}{lcccc}
\toprule
 Methods & Type & PSNR$\uparrow$  & SSIM$\uparrow$ & LPIPS$\downarrow$ \\
\cmidrule{1-1} \cmidrule(lr){2-2}  \cmidrule(lr){3-5}
T-NeRF\,\cite{pumarola2021_dnerf_cvpr21} & N & 29.51 & 0.95 & 0.08 \\
\midrule
 TiNeuVox-S ($100^3$)\,\cite{fang2022_TANV_arxiv} & NPC
    & 30.75 & \tabsecond{0.96} & 0.07 \\
TiNeuVox-B ($160^3$)\,\cite{fang2022_TANV_arxiv}  & NPC
    & \tabsecond{32.67} & \tabfirst{0.97} & \tabfirst{0.04} \\
\midrule
D-NeRF\,\cite{pumarola2021_dnerf_cvpr21} & PC
    & 30.50 & 0.95 & 0.07 \\
NDVG ($160^3$)\,\cite{guo2022_NDVG_arxiv} & PC
     & 30.54 & \tabsecond{0.96} & \tabsecond{0.05} \\
\hdashline
Ours ($80^3$)\, & PC
    & \tabfirst{32.68} & \tabfirst{0.97} & \tabfirst{0.04} \\
\bottomrule
\end{tabular}
}
\vspace{-\baselineskip}

\end{table}

\subsection{Baselines and Evaluation Datasets}
\paragraph{Baselines} 
The methods compared in this paper are classified into three types.
First, we compare methods which are \emph{non-canonical based} (\textbf{N}), including NeRF\cite{mildenhall2020_nerf_eccv20}, and T-NeRF\cite{pumarola2021_dnerf_cvpr21}. Second, \emph{physical canonical based methods} (\textbf{PC}) contain D-NeRF \cite{pumarola2021_dnerf_cvpr21}, NDVG\cite{guo2022_NDVG_arxiv} and Ours. These methods set their canonical space as one frame of the whole timeline by explicitly giving zeros to estimated flows at the canonical time. The canonical space of these methods should have reasonable physical 3D reconstructions of the scene at the canonical time. Third, \emph{non-physical canonical based methods} (\textbf{NPC}) consists of NV\cite{Lombardi2019_NV_ACMTG}, Nerfies\cite{park2021_nerfies_iccv21}, HyperNeRF\cite{park2021hypernerf} and TiNeuVox\cite{fang2022_TANV_arxiv}. Their canonical space geometries do not necessarily have physical meanings.

\paragraph{D-NeRF Dataset} 
We evaluate our method on 8 dynamic scenes of D-NeRF~\cite{pumarola2021_dnerf_cvpr21} dataset. 
We report several common metrics for the evaluation: Peak Signal-to-Noise Ratio (PSNR), Structural Similarity (SSIM) and Learned Perceptual Image Patch Similarity (LPIPS)~\cite{zhang2018_lpips_CVPR}.

\paragraph{HyerNeRF Dataset}
Besides synthetic datasets, we also conduct experiments on real scenes, proposed by HyperNeRF \cite{park2021hypernerf}. This dataset uses a multi-view camera rig consisting of 2 phones to capture real unbounded scenes with challenging rigid and non-rigid deformations.

\paragraph{NHR Dataset}
NHR~\cite{wu2020_NHR} dataset capture dynamic human data using a multi-camera dome system with up to 80 cameras arranged on a cylinder. We conduct experiments on four scenes where the performers are in different clothing and perform different actions.  We test with 100 frames of each scene, selecting 90\% views for training and 10\% views for testing.

\paragraph{Lego Complete Dataset} The D-NeRF~\cite{pumarola2021_dnerf_cvpr21} dataset only contains 20 test images for each scene, and the evaluations of D-NeRF do not contain images in the canonical space. However, for \emph{physical canonical based methods} (\textbf{PC}), canonical space geometry is important, as it affects images quality rendered at the canonical time and reflects the quality of flow estimation.
Hence, we build a new dataset, named \emph{Lego Complete Dataset}, which animates the object \emph{LEGO} in D-NeRF dataset with 3 different motion patterns. For each scene, the test set is split into 3 categories to evaluate three abilities: \emph{space interpolation} (rand views for each train time step), \emph{time interpolation}(times between trained time steps), and \emph{canonical interpolation} (rand views for canonical time step) abilities. For more details, please refer to our \textit{supp}. %

Besides the image quality, we also consider geometry precision by evaluating the depth of each test image to analyze the reconstructed geometry. For depth metrics, we use the relative error
    $\delta = \text{max}\left(\frac{d}{d_{\text{gt}}}, \frac{d_{\text{gt}}}{d}\right),$
where $d$ is the estimated depth and $d_{\text{gt}}$ is the ground truth depth. We report the percentage of the pixels with $\delta < \text{Threshold}$.

\begin{figure}[!t] \centering
    \includegraphics[width=\columnwidth]{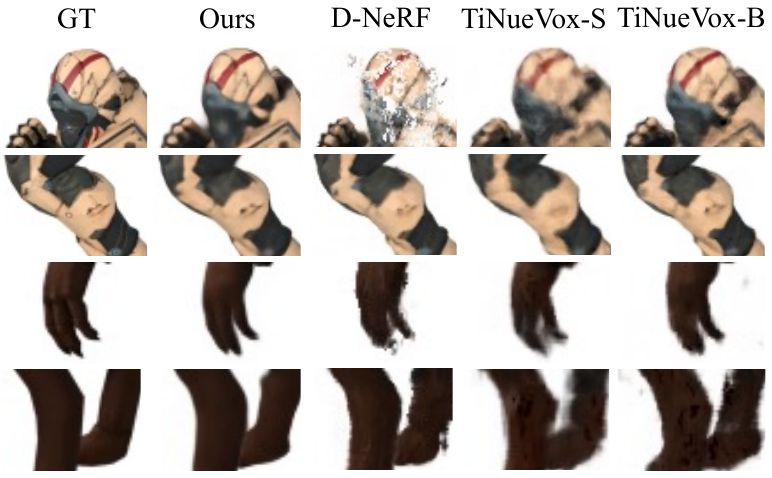}
    \vspace{-\baselineskip}
    \caption{\textbf{Qualitative comparison on D-NeRF Dataset.} We show some novel view synthesized images on the selected test set of the dataset. Comparing ours with ground truth, D-NeRF~\cite{pumarola2021_dnerf_cvpr21} and TiNeuVox~\cite{fang2022_TANV_arxiv}. Our model yields cleaner images with more details.
    } \label{fig:viscompare}
    \vspace{-2\baselineskip}
\end{figure}

\begin{table*}[t!]
\setlength{\tabcolsep}{3pt} %
\centering
\caption{\small{
\textbf{Quantitative comparison on real scene HyperNeRF dataset.} 
Comparison of our method with others on PSNR and MS-SSIM. 
}}
\label{table:hyperquant}
\vspace{-0.5\baselineskip}
\footnotesize{
\begin{tabular}{lccccccccccc}
\toprule
& & \multicolumn{2}{c}{3D Printer} & \multicolumn{2}{c}{Broom} & \multicolumn{2}{c}{Chicken} & \multicolumn{2}{c}{Peel Banana} & \multicolumn{2}{c}{Mean}\\
Methods & Type
& PSNR$\uparrow$  & MS-SSIM$\uparrow$
& PSNR$\uparrow$  & MS-SSIM$\uparrow$
& PSNR$\uparrow$  & MS-SSIM$\uparrow$
& PSNR$\uparrow$  & MS-SSIM$\uparrow$
& PSNR$\uparrow$  & MS-SSIM$\uparrow$\\
\cmidrule{1-1} \cmidrule(lr){2-2} \cmidrule(lr){3-4} \cmidrule(lr){5-6} \cmidrule(lr){7-8} \cmidrule(lr){9-10}  \cmidrule(lr){11-12}

NeRF~\cite{mildenhall2020_nerf_eccv20} & N 
    & 20.7 & 0.780 
    & 19.9 & 0.653 
    & 19.9 & 0.777 
    & 20.0 & 0.769 
    & 20.1 & 0.745 \\
\midrule
NV~\cite{Lombardi2019_NV_ACMTG} & NPC
    & 16.2 & 0.665 
    & 17.7 & 0.623 
    & 17.6 & 0.615 
    & 15.9 & 0.380 
    & 16.9 & 0.571 \\

Nerfies~\cite{park2021_nerfies_iccv21} & NPC
    & 20.6 & 0.830 
    & 19.2 & 0.567 
    & 26.7 & 0.943 
    & 22.4 & 0.872 
    & 22.2 & 0.803 \\
    
HyperNeRF~\cite{park2021hypernerf} & NPC
    & 20.0 & 0.821 
    & 19.3 & 0.591 
    & 26.9 & \tabfirst{0.948} 
    & 23.3 & \tabfirst{0.896} 
    & 22.4 & 0.814 \\

TiNeuVox-S ($100^3$)~\cite{fang2022_TANV_arxiv} & NPC
    & \tabsecond{22.7} & 0.836
    & \tabfirst{21.9} & \tabsecond{0.707}
    & 27.0 & 0.929
    & 22.1 & 0.780
    & 23.4 & 0.813 \\
    
TiNeuVox-B ($160^3$)~\cite{fang2022_TANV_arxiv} & NPC
    & \tabfirst{22.8} & \tabsecond{0.841} 
    & \tabsecond{21.5} & 0.686 
    & \tabfirst{28.3} & \tabsecond{0.947}
    & \tabfirst{24.4} & \tabsecond{0.873}
    & \tabfirst{24.3} & \tabsecond{0.837} \\

\midrule

NDVG ($160^3$)~\cite{guo2022_NDVG_arxiv} & PC
    & 22.4 & 0.839 
    & \tabsecond{21.5} & 0.703 
    & 27.1 & 0.939 
    & 22.8 & 0.828  
    & 23.3 & 0.823 \\

\hdashline
Ours ($70^3$)\, & PC
    & \tabfirst{22.8} & \tabfirst{0.845}
    & \tabfirst{21.9} & \tabfirst{0.715}
    & \tabsecond{28.0} & 0.944
    & \tabsecond{24.3} & 0.865 
    & \tabsecond{24.2} & \tabfirst{0.842} \\

\bottomrule %

\end{tabular}
}
\vspace{-\baselineskip}
\end{table*}

\begin{figure*}[!t] \centering
    \includegraphics[width=\linewidth]{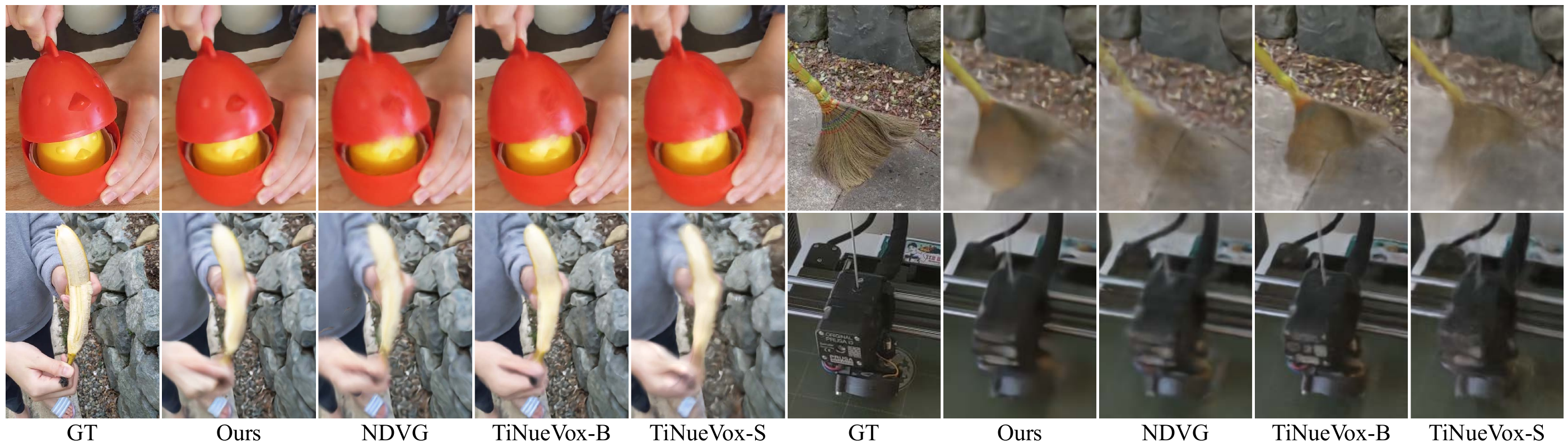}
    \vspace{-1.5\baselineskip}
    \caption{\textbf{Qualitative comparison on HyperNeRF Dataset.} 
    Our results are closer to ground truth than other methods.
    } \label{fig:hypercompare}
    \vspace{-1.5\baselineskip}
\end{figure*}

\subsection{Experimental Results}
\paragraph{Evaluation on D-NeRF Dataset}
We compared our method with the canonical-based backward flow methods D-NeRF~\cite{pumarola2021_dnerf_cvpr21}, NDVG~\cite{guo2022_NDVG_arxiv} and TiNeuVox~\cite{fang2022_TANV_arxiv} on D-NeRF Dataset. As shown in~\Tref{table:quant}, our method achieves significant improvements in terms of all three metrics for \emph{physical canonical based methods}. On average, our method improves PSNR by 2.18 compared with D-NeRF and 2.14 compared with NDVG. For \emph{non-physical canonical based methods}, our method also achieves the best result. Ours show obviously better qualitative rendering quality compared with TiNeuVox-S ($100^3$)~\cite{fang2022_TANV_arxiv} with even smaller voxel resolution. More detailed statistics are provided in \textit{supp}. %

We also provide some visual comparisons in~\fref{fig:viscompare}. Ours could recover accurate and detailed images, \eg, the helmet and arm in the top scene, and could also produce cleaner boundaries, \eg the hand and feet in the bottom scene.

\paragraph{Evaluation on HyperNeRF Dataset}
We further compare our method with some highly related works on real scene dataset proposed by~\cite{park2021hypernerf}. As shown in~\Tref{table:hyperquant}, our method achieves consistently better performance among \emph{physical canonical based methods}. Our method has a voxel grid resolution limitation, due to we need to warp the whole voxel grid for rendering. Compared with other voxel grid based methods, our voxel grid resolution is limited to $70^3$, along with a $160^3$ static voxel grid for background modeling. However, the performance of our method is still comparable with SOTA of \emph{non-physical canonical based methods}, proving the effectiveness of our forward flow design.

In \fref{fig:hypercompare}, we show visual comparisons with NDVG \cite{guo2022_NDVG_arxiv}, and TiNueVox \cite{fang2022_TANV_arxiv}. The forward flow helps to recover the correct structure of dynamic objects, like the eye and edges of the chicken (top left) and the broom (top right). The relatively low resolution may harm the ability to recover very fine details, like the patterns on the 3D printer compared with TiNueVox-B \cite{fang2022_TANV_arxiv} (bottom right).

\begin{table*}[t!]
\setlength{\tabcolsep}{4pt} %
\centering
\caption{\small{
\textbf{Quantitative comparison on NHR dataset.} The \tabfirst{red} text indicates the best and \tabsecond{blue} text is the second best result. }}
\label{table:quantnhrdetail}
\resizebox{\textwidth}{!}{%
\begin{tabular}{lcccccccccccccccc}
\toprule
& & \multicolumn{3}{c}{Sport 1} & \multicolumn{3}{c}{Sport 2} & \multicolumn{3}{c}{Sport 3} & \multicolumn{3}{c}{Bacsketball} & \multicolumn{3}{c}{Mean} \\
Methods & type
& PSNR$\uparrow$  & SSIM$\uparrow$ & LPIPS$\downarrow$
& PSNR$\uparrow$  & SSIM$\uparrow$ & LPIPS$\downarrow$
& PSNR$\uparrow$  & SSIM$\uparrow$ & LPIPS$\downarrow$
& PSNR$\uparrow$  & SSIM$\uparrow$ & LPIPS$\downarrow$
& PSNR$\uparrow$  & SSIM$\uparrow$ & LPIPS$\downarrow$\\
\cmidrule{1-1} \cmidrule(lr){2-2} \cmidrule(lr){3-5} \cmidrule(lr){6-8} \cmidrule(lr){9-11} \cmidrule(lr){12-14} \cmidrule(lr){15-17}
    TiNeuVox-S \,\cite{fang2022_TANV_arxiv} & NPC
    & 26.06 & \tabsecond{0.93} & \tabsecond{0.10}
    & 25.98 & 0.93 & 0.11
    & 25.90 & \tabsecond{0.93} & \tabsecond{0.11}
    & 23.75 & 0.91 & 0.14
    & 25.42 & 0.92 & 0.12 \\
    TiNeuVox-B \,\cite{fang2022_TANV_arxiv} & NPC
    & \tabsecond{26.44} & \tabsecond{0.93} & \tabsecond{0.10}
    & \tabsecond{26.68} & \tabsecond{0.94} & \tabsecond{0.10}
    & \tabsecond{26.09} & \tabsecond{0.93} & \tabsecond{0.11}
    & \tabfirst{25.06} & \tabsecond{0.92} & \tabsecond{0.12}
    & \tabsecond{26.07} & \tabsecond{0.93} & \tabsecond{0.11} \\
    \midrule
    NDVG\,\cite{guo2022_NDVG_arxiv} & PC
    & 23.66 & 0.89 & 0.15
    & 24.43 & 0.91 & 0.13
    & 22.54 & 0.88 & 0.16
    & 22.55 & 0.89 & 0.17 
    & 23.29 & 0.89 & 0.15 \\
    
    \hdashline
    
    Ours\,  & PC
    & \tabfirst{27.71} & \tabfirst{0.95} & \tabfirst{0.08}
    & \tabfirst{27.89} & \tabfirst{0.95} & \tabfirst{0.08}
    & \tabfirst{27.57} & \tabfirst{0.94} & \tabfirst{0.08}
    & \tabsecond{24.85} & \tabfirst{0.93} & \tabfirst{0.11}
    & \tabfirst{27.00} & \tabfirst{0.94} & \tabfirst{0.09} \\

\bottomrule
\end{tabular}
}
\vspace{-0.5\baselineskip}

\end{table*}
\paragraph{Evaluation on NHR Dataset}
We also test our method on NHR dataset. Quantitative results are shown in \Tref{table:quantnhrdetail}, which shows ours achieving best results consistently. We show qualitative comparison in \fref{fig:nhrvistraj}, and our method could render clean and detailed images.

\begin{table*}[!t]
\setlength{\tabcolsep}{3pt} %
\centering
\caption{\small{\textbf{Quantitative comparison on Lego Complete Dataset.} We report the average PSNR, LPIPS and the average relative depth error.}}

\label{table:quantcomplete}
\vspace{-0.5\baselineskip}
\footnotesize{%
\begin{tabular}{lccccccccccccc}
\toprule
& & \multicolumn{4}{c}{Space Interpolation} & \multicolumn{4}{c}{Time  Interpolation} & \multicolumn{4}{c}{Canonical  Interpolation} \\
Methods & Type
& PSNR$\uparrow$ & LPIPS$\downarrow$ & $\delta\!<\!1.25$ & $\delta\!<\!1.25^2$
& PSNR$\uparrow$ & LPIPS$\downarrow$ & $\delta\!<\!1.25$ & $\delta\!<\!1.25^2$
& PSNR$\uparrow$ & LPIPS$\downarrow$ & $\delta\!<\!1.25$ & $\delta\!<\!1.25^2$\\
\cmidrule{1-1} \cmidrule(lr){2-2}  \cmidrule(lr){3-6} \cmidrule(lr){7-10} \cmidrule(lr){11-14}

    D-NeRF\,~\cite{pumarola2021_dnerf_cvpr21} & PC
    & 23.98 & 0.15 & 98.720 & 99.328
    & 24.18 & 0.15 & 98.854 & 99.378
    & 17.36 & 0.23 & 97.341 & 98.868 \\
    NDVG ($160^3$)~\cite{guo2022_NDVG_arxiv}  & PC
    & 28.12 & \textbf{0.06} & 99.655 & 99.853
    & 28.19 & \textbf{0.06} & 99.749 & 99.953
    & 19.90 & 0.11 & 94.828 & 97.560 \\

\midrule
    Ours\_notv ($80^3$)\,$^1$  & PC
    & 27.58 & 0.07 & 99.758 & 99.958
    & 27.63 & 0.07 & 99.788 & 99.966
    & 23.56 & 0.13 & 99.238 & 99.820 \\
    Ours\_noup ($80^3$)\,$^2$  & PC
    & 26.11 & 0.09 & 99.596 & 99.981
    & 26.19 & 0.09 & 99.636 & 99.983
    & 24.64 & 0.10 & 99.230 & 99.925 \\
    Ours\_noinp ($80^3$)\,$^3$  & PC
    & 27.21 & 0.08 & 99.783 & 99.979
    & 27.27 & 0.08 & 99.797 & 99.980
    & 23.91 & 0.09 & 99.586 & 99.963 \\
    Ours ($80^3$)\,  & PC
    & \textbf{28.18} & \textbf{0.06} & \textbf{99.818} & \textbf{99.985}
    & \textbf{28.22} & \textbf{0.06} & \textbf{99.839} & \textbf{99.987}
    & \textbf{25.63} & \textbf{0.07} & \textbf{99.654} & \textbf{99.968} \\

\bottomrule %

\multicolumn{10}{l}{$^1$~\footnotesize{not use the all total variation losses} \quad $^2$~\footnotesize{not up-sample the voxel grid} \quad $^3$~\footnotesize{not use the inpaint network}}

\end{tabular}

}
\vspace{-1.5\baselineskip}

\end{table*}

\begin{figure}
    \centering
    \vspace*{-0.5\baselineskip}
    \captionsetup{type=figure}
    \subfloat{\includegraphics[trim={0 0 0 20},clip,width=0.40\linewidth]{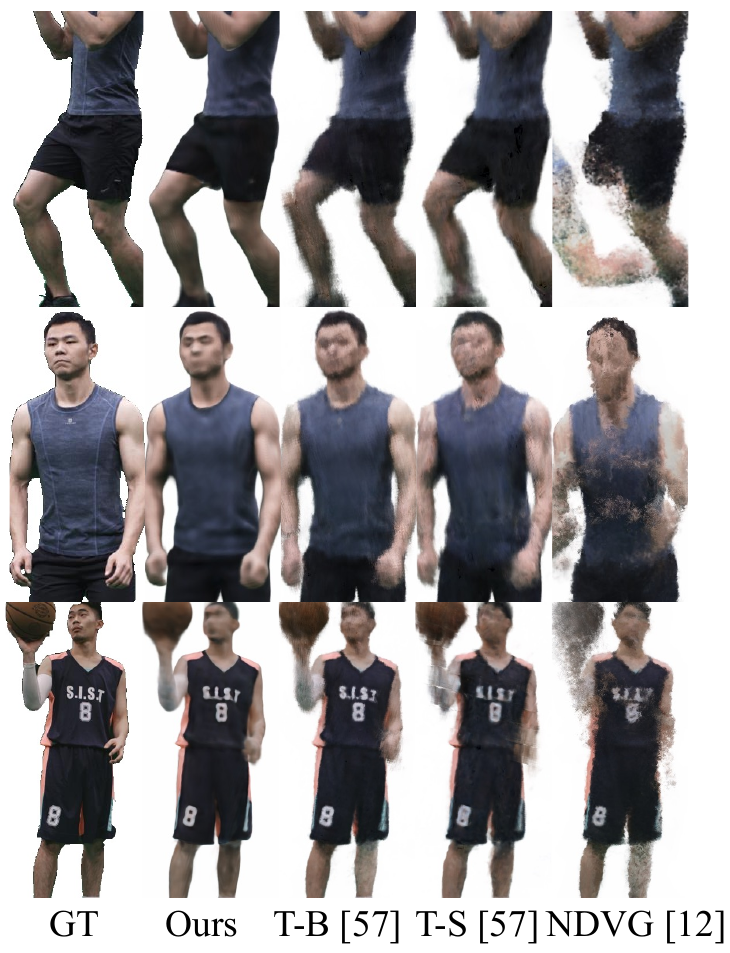}}
    \hspace{0.01mm}
    \subfloat{\includegraphics[trim={70 -30 170 1},clip,width=0.32\linewidth]{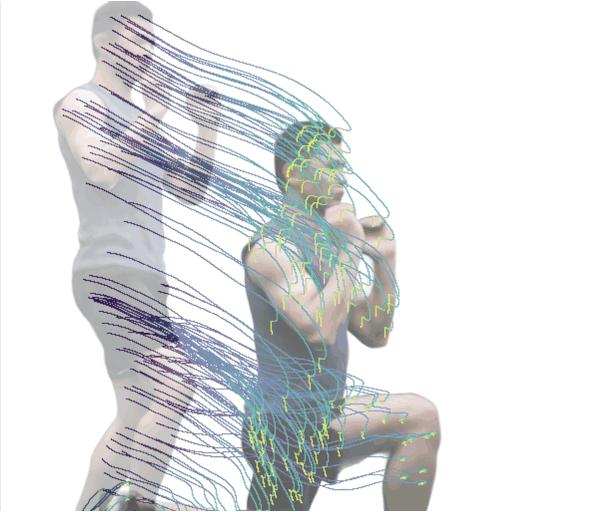}}
    \hspace{0.01mm}
    \subfloat{\includegraphics[trim={250 -30 100 40},clip,width=0.26\linewidth]{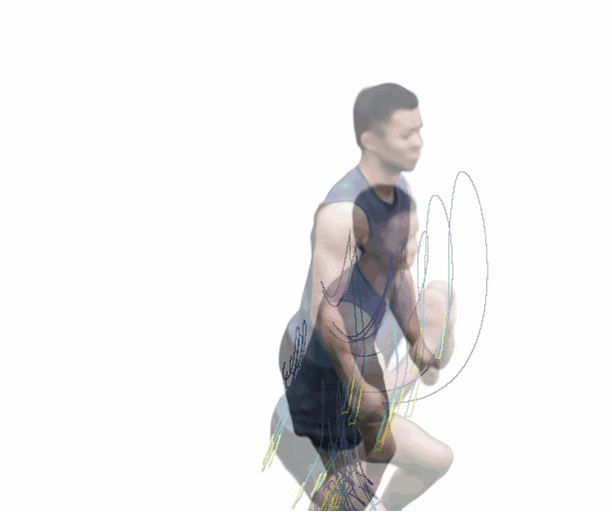}}

    \captionof{figure}{\footnotesize{\textbf{Visual comparisons on NHR dataset.}}
    \label{fig:nhrvistraj}}
    \vspace*{-1.8\baselineskip}
\end{figure}%

\subsection{Method Analysis}

\paragraph{Flow Estimation and Canonical Space Geometry}
Since evaluating the estimated flow with ground truth is not practical, we choose to compare the reconstructed canonical geometry of \emph{physical canonical based methods}. The idea is that if the canonical space is one frame of the whole timeline (like D-NeRF \cite{pumarola2021_dnerf_cvpr21} and NDVG \cite{guo2022_NDVG_arxiv}), better flow estimation will result in better canonical scene geometry. We test these methods on Lego Complete Dataset, and all set the canonical space as the first frame of the image sequence. 

In~\Tref{table:quantcomplete}, D-NeRF can not reconstruct the canonical radiance field well, since both the image and depth quality in canonical interpolation are significantly worse than those in space interpolation and time interpolation. NDVG \cite{guo2022_NDVG_arxiv} could recover much better images and geometry for other time steps, but the differences between canonical interpolation with the other two are even more significant. This means NDVG \cite{guo2022_NDVG_arxiv} has difficulty estimating correct backward flows. To render good quality images at other time steps, the correct geometries (depth) at these time steps are warped back into distorted canonical geometry due to incorrect backward flows.

This is the proof of our claim that the non-smooth and discontinuous nature of the backward flow deformation field makes it difficult to fit with smooth functions, especially for NDVG \cite{guo2022_NDVG_arxiv}, which has limited MLP capacity for fast training.
However, our method performs significantly better for both image quality and depth, and there are fewer variations between canonical, space, and temporal interpolations. This demonstrates the benefit of forward warping, which makes it easier for the deformation model to learn deformation flows from canonical time to other times.

\begin{figure}[!t] \centering
    \includegraphics[width=0.9\columnwidth]{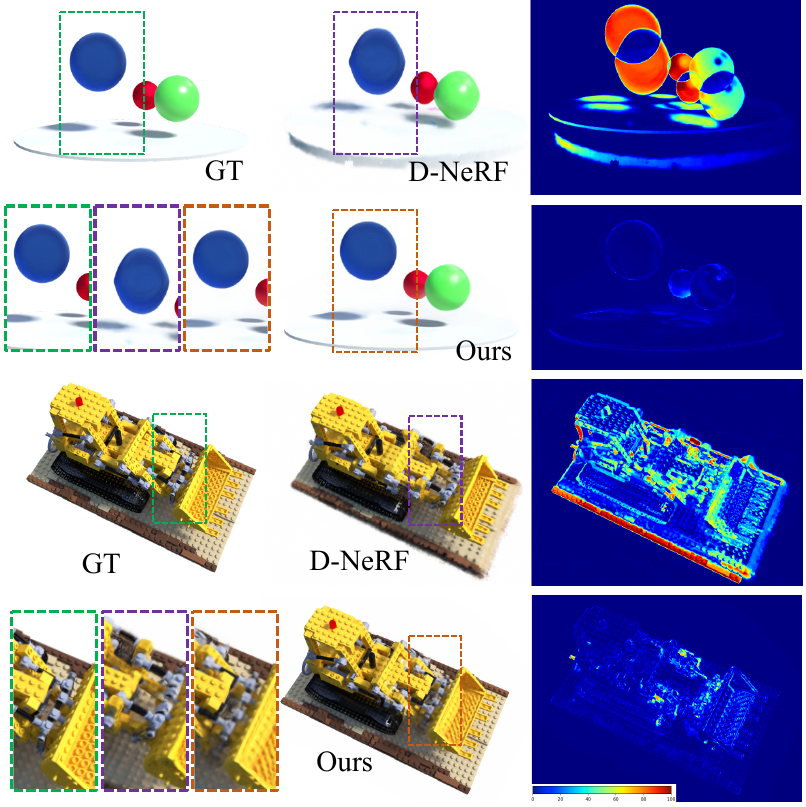}
    \vspace{-0.5\baselineskip}
    \caption{\textbf{Canonical qualitative comparison.} We show canonical radiance field comparison with D-NeRF~\cite{pumarola2021_dnerf_cvpr21}.
    Given the error map between the ground truth and rendered images, we can see that the canonical frame yielded by ours is closer to the ground truth. The results of D-NeRF are blurry and have large displacements.
    } \label{fig:cancompare}
    \vspace{-2.5\baselineskip}
\end{figure}

We compare the canonical image of our method with D-NeRF \cite{pumarola2021_dnerf_cvpr21} in \fref{fig:cancompare}. Our method could recover the correct geometry and image details of the canonical frame. For example, the arms of the bucket produced by D-NeRF are at the ``mean'' positions of the whole trajectory, and ours recover the correct status.

\begin{figure}[!t] \centering
    \vspace{-0.5\baselineskip}
    \includegraphics[width=0.95\columnwidth]{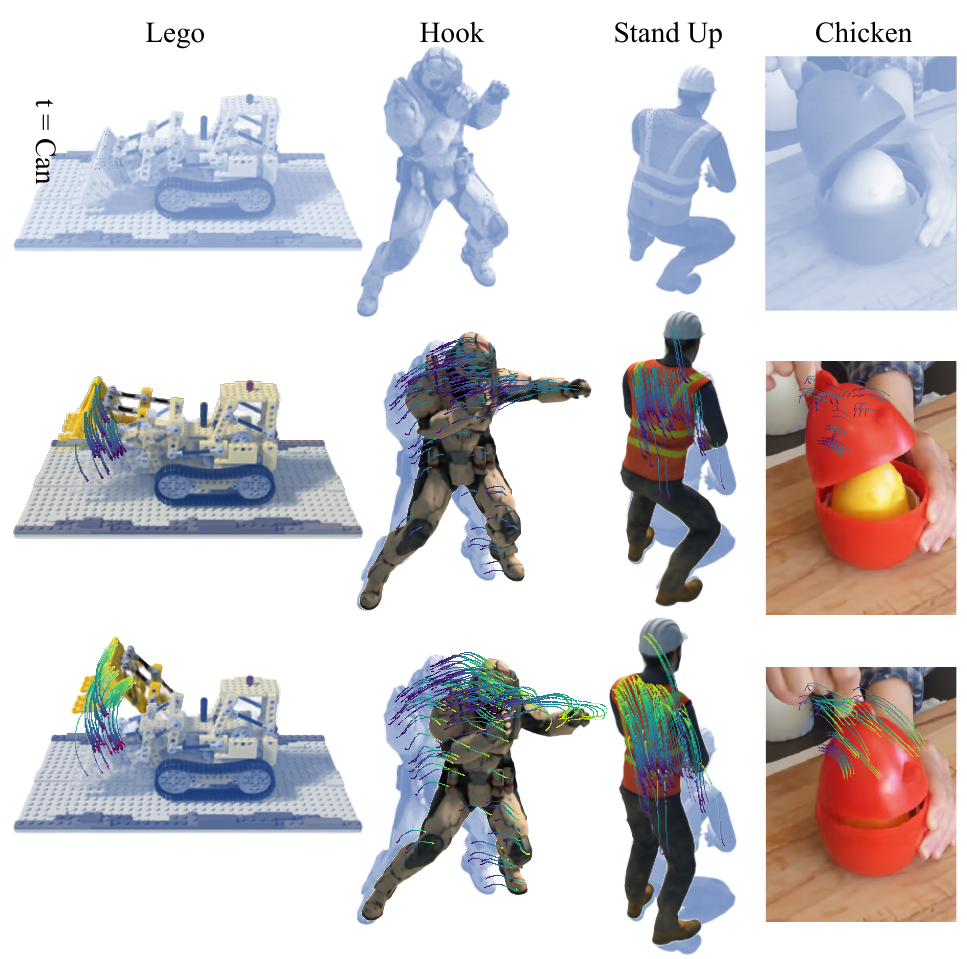} %
    \vspace{-1.0\baselineskip}
    \caption{\textbf{Trajectory learned by the canonical trajectory field.} Light blue is the canonical frame, the curve represents the historical motion trajectory. More videos can be found in \textit{supp}.} 

    \label{fig:vistraj}
    \vspace{-2.5\baselineskip}
\end{figure}

\paragraph{Trajectory Visualization}
We show the learned DCT trajectories generated by our canonical trajectory field in~\fref{fig:nhrvistraj} and~\fref{fig:vistraj}. 
The canonical trajectory field is capable of recovering reasonable object point trajectories.
Backward warping designs cannot track the motions of the same object point, so this is not a viable solution. 
With this feature, geometry constraints, motion models, and prior knowledge could be introduced in future work.

\paragraph{Regularization terms} 
We study the effect of all regularization terms we proposed, including $\Loss{flow}$, $\Loss{vdiff}$ and $\Loss{tv}$ in \Tref{tab:rebuttalmain}. We could observe that all these three regularization terms has positive effect on the performance, but the improvement is minor. This proves the improvement of our method compared with others come from the forward warping desgin we proposed in the paper. Also, backward flow based method NDVG \cite{guo2022_NDVG_arxiv} use similar regularization terms with ours, and our method has clear advantage compared with NDVG \cite{guo2022_NDVG_arxiv}.

\paragraph{Canonical setup} 
In our method, we set canonical time to be the first frame for D-NeRF dataset to compare with \emph{physical canonical based method} and the middle frame for HyperNeRF dataset for better performance. Setting canonical time to be the middle frame helps improving the performance as the state of the middle frame geometry is closer to other time steps compared with the first frame. To prove this, we set canoical time to be middel frame in \Tref{tab:rebuttalmain} (can-time t at mid), and the PSNR improves slightly.

\paragraph{Photmetric loss for $\Vgrid{\text{R}_{\text{Up}}}{}$} 
We use photometric terms on $\Vgrid{\text{R}_{\text{Up}}}{}$ to make sure the warped grid before inpainting could already render reasonable images. This make sure the UNet is actually doing `inpainting'. Also, we do observe inpainting and upsampling could refine grids. In~\fref{fig:nhrvistraj} and~\fref{fig:vistraj}, the trajectories are reasonable which means we do learn trajectories without inpainting overfitting. We also test w/o photometric terms of $\Vgrid{\text{R}_{\text{Up}}}{}$ in \Tref{tab:rebuttalmain} (w/o $\Vgrid{\text{R}_{\text{Up}}}{}$ photo) and the performance drops as there is no direct supervise signal for trajectory learning.

\begin{table}[!t]
\centering
\caption{
\textbf{Ablations.} Mean of Hell Warrior, Mutant, Hook and Bouncing Balls in D-NeRF dataset.
}
\setlength\tabcolsep{3pt}
\label{tab:rebuttalmain}
    \resizebox{0.9\columnwidth}{!}{
    \large
    \begin{tabular}{*{9}{c}}
        \toprule
       Method & Ours &  w/o            & w/o           &       w/o       &  can-time  & w/o $\Vgrid{\text{R}_{\text{Up}}}{}$\\
              &      &   $\Loss{tv}$   & $\Loss{flow}$ & $\Loss{vdiff}$  &  t at mid  &      photo                           \\

        \midrule
        PSNR  & 33.75 & 33.30 & 33.74 & 33.57 & 34.09 & 31.56 \\
        SSIM  & 0.979 & 0.974 & 0.978 & 0.977 & 0.980 & 0.966 \\
        LPIPS & 0.040 & 0.048 & 0.039 & 0.041 & 0.037 & 0.057 \\
        \bottomrule
    \end{tabular}
    }
    \vspace{-2.0em}
\end{table}

\begin{table}[t!]
\setlength{\tabcolsep}{4pt} %
\centering
\caption{\small{\textbf{Ablation of losses on two dataset. }}}
\vspace{-0.5\baselineskip}
\label{table:quantablation}
\footnotesize{%
\begin{tabular}{lcccc}
\toprule
& \multicolumn{2}{c}{D-NeRF} & \multicolumn{2}{c}{HyperNeRF} \\
Methods 
& PSNR$\uparrow$  & LPIPS$\downarrow$ & PSNR$\uparrow$  & MS-SSIM$\uparrow$ \\
\cmidrule{1-1}  \cmidrule(lr){2-3} \cmidrule(lr){4-5}
    Ours\_notv\,$^1$
        & 33.09 & 0.046 & 23.84 & 0.831 \\
    Ours\_noup\,$^2$
        & 32.03 & 0.043 & 23.72 & 0.822 \\
    Ours\_noinp\,$^3$
        & 31.78 & 0.044 & 23.25 & 0.814 \\
\midrule
    Ours \,
    & 32.68 & 0.037 & 24.23 & 0.842 \\

\bottomrule %

\end{tabular}
}
\vspace{-2.0\baselineskip}

\end{table}

\paragraph{More Ablations}
We study the effects of the total variation losses, the up-sample strategy, and the inpaint network we proposed in \Tref{table:quantablation}. The total variation losses help to improve visual quality (LPIPS) in D-NeRF dataset as it applies smoothness over the deformation and radiance space, but may harm the high-frequency details (PSNR) for some of the scenes in D-NeRF dataset. For HyperNeRF dataset, the total variation losses improve the results, as the camera settings are more challenging than D-NeRF dataset. However, this improvement is minor compared with the other two factors. Since the resolution is critical for voxel grid based method, up-sampling operation of inpaint network is vital to improve the performance. Finally, the inpaint network plays the most essential role among these three factors in our method. More ablations are provided in our \textit{supp}.
\section{Conclusion}\label{sec:conclusion}

This paper presents a canonical-based representation with forward warping for novel view synthesis of dynamic scenes. Our method models a static scene at the canonical field and forward warp this whole field to other time steps for dynamic scene rendering. 
To address the many-to-one and one-to-many mapping difficulties, we present a differentiable forward warping with an average splatting process and an inpaint network.
Our proposed forward warping pipeline achieves SOTA performance on the public datasets and our newly built dataset, especially for canonical frames.

\paragraph{Limitations and Future Directions} Our current implementation is relatively memory-consuming, especially for real scenes. Also, the training speed is relatively slow (one day for each scene). 
Since we have a smooth trajectory field thanks to the forward warping, additional constraints and motion models could be introduced to learn better trajectories in future works.

\paragraph{Acknowledgements}
  This research was supported in part by the National Natural Science Foundation of China (Grant Nos.~62271410, 62293482, 62202409), the Fundamental Research Funds for the Central Universities, and Shenzhen Science and Technology Program (Grant No. RCBS20221008093241052).

\appendix

\section{More Implementation Details}

\subsection{Preliminaries}

\begin{table}[t!]
\setlength{\tabcolsep}{4pt} %
\centering
\caption{\small{
\textbf{Preliminaries in the paper.}
}}
\label{table:preliminaries}
\footnotesize{%
\begin{tabular}{ll}
\toprule

Variables & Description  \\
\cmidrule{1-1} \cmidrule{2-2} 
$\Vgrid{\text{Rf}}{\text{Can}}$ & voxel grid contains canonical radiance feature \\
$\Vgrid{\text{p}}{\text{Can}}$ & voxel grid contains coordinate of each voxel \\
$F_{\theta_1}$ & light weight MLP network estimating canonical radiance field \\
$\Vgrid{\text{R}}{\text{Can}}$ & canonical radiance field estimated by $F_{\theta_1}$ \\
$\Vgrid{\sigma}{\text{Can}}$ & canonical density voxel grid in $\Vgrid{\text{R}}{\text{Can}}$ \\
$\Vgrid{\text{cf}}{\text{Can}}$ & canonical color feature voxel grid in $\Vgrid{\text{R}}{\text{Can}}$ \\
$\Vgrid{\text{Df}}{\text{Can}}$ & voxel grid contains canonical deformation feature \\
$F_{\theta_2}$ & light weight MLP network estimating trajectory of each voxel \\
$\Vgrid{T}{\text{Can}}$ & trajectory voxel grid, containing DCT params of each voxel \\
$\Vgrid{\text{flow}}{t}$ & deformation flow of each voxel from canonical to time $t$ \\
$\Warper$ & average splatting operation \\
$\Vgrid{\text{R}}{t}$ & radiance field at time $t$ warped from $\Vgrid{\text{R}}{\text{Can}}$ \\
$\InNet$ & Inpaint Network \\
$\Vgrid{\text{R}_{\text{Inp}}}{t}$ & inpainted radiance field at time $t$ by $\InNet$ \\
$\Vgrid{\text{R}_{\text{Up}}}{t}$ & upsampled radiance field at time $t$ \\
$\Renderer$ & volume rendering function \\
$\mathbf{C_{\text{Inp}}}(\mathbf{r})$ & color of ray $r$ rendered from field $\Vgrid{\text{R}_{\text{Inp}}}{t}$ \\
$\mathbf{C_{\text{Up}}}(\mathbf{r})$ & color of ray $r$ rendered from field $\Vgrid{\text{R}_{\text{Up}}}{t}$ \\

\bottomrule
\end{tabular}
}

\end{table}

We provide preliminaries in \Tref{table:preliminaries}

\subsection{Network Architecture}
There are four networks in proposed network: canonical radiance network $\RadNet$, canonical deformation network $\DefNet$, view dependent color network $\ViewNet$ and 
inpaint network $\InNet$.

Canonical radiance network $\RadNet$ is a 3 layer MLP, with width set to 128. The input contains the radiance feature sampled from radiance feature grid $\Vgrid{\text{Rf}}{\text{Can}}$ with dimension 12, and embedded position with dimension 33. The output contains density whose dimension is 1, and color feature whose dimension is 3 or 12, depending the training stage.

Canonical deformation network $\DefNet$ is a 4 layer MLP, with width set to 64. Similar with canonical radiance network,  the input of $\DefNet$ contains the deformation feature and embedded position with dimension 33. We set the deformation feature dimension to be 0 and the number of dct bases to be 15 for D-NeRF dataset. For Hypernerf dataset, we set the deformation feature dimension to be 12 and the number of dct bases to be 25.

View dependent color network $\ViewNet$ is a simpler MLP with only 2 layers and width is 64. For input, the dimension of embedded view direction is 27 and color feature is 12. The output is the rgb color with dimension 3.

Inpaint network $\InNet$ consist of a UNet structure and a up-sample layer. The UNet structure has 4 encode layers and 3 decode layers. Each encode layer consists of one max pooling layer and two convolution layers. For one convolution layer, there is one instance norm, followed by convolution and ReLU activation. Each decode layer consists of one up-sample interpolation layer and two convolution layers, which are the same with encode layer. The up-sample layer has the same structure with the decode layer.

\subsection{Average Splatting vs. Softmax Splatting}

We use average splatting in this paper, rather then more complex splatting methods in \cite{Niklaus2020_VFI_CVPR}, like softmax splatting. We tried to predict weights for softmax splatting, and there is no obvious improvement compared with average splatting. It may introduce too much complexity and we find the UNet could refine the voxel grid to some extend. Average splatting is enough in our setting.

\subsection{Losses and Hyper-parameter Settings}
Following DVGO~\cite{sun2021direct}, we use $\Loss{ptc}$ to directly supervise the color of sampled points. The intuition is that sampled points with bigger weights contribute more to the rendered color.
\begin{equation}
    \Loss{ptc} =\frac{1}{|\mathcal{R}|} \sum_{r\in\mathcal{R}} \Loss{}(\mathbf{C_{\text{Inp}}}(\mathbf{r})) + \frac{1}{|\mathcal{R}|} \sum_{r\in\mathcal{R}} \Loss{}(\mathbf{C_{\text{Up}}}(\mathbf{r})),
\end{equation}

\begin{equation}
    \Loss{}(\mathbf{C}(\mathbf{r})) =  \frac{1}{K}
        \sum\nolimits_{k=1}^{K} A_{\text{accum}}(k) \left\|\mathbf{c}(w_k) - \mathbf{C}_{\text{gt}}(\mathbf{r}) \right\|_2^2 ,
\end{equation}

\begin{equation}
    A_{\text{accum}}(k) =   T(w_k) \, \decay{\sigma(w_k) \delta_{k}} .
\end{equation}

 Also, we use the background entropy loss $\Loss{bg}$ to encourage the densities concentrating on either the foreground or the background. 
 
\begin{equation}
    \Loss{bg} =\frac{1}{|\mathcal{R}|} \sum_{r\in\mathcal{R}} \Loss{}(\mathbf{A_{\text{Inp}}}(\mathbf{r})) + \frac{1}{|\mathcal{R}|} \sum_{r\in\mathcal{R}} \Loss{}(\mathbf{A_{\text{Up}}}(\mathbf{r})),
\end{equation}

\begin{gather}
\begin{split}
    \Loss{}(\mathbf{A}(\mathbf{r})) =& -\frac{1}{K-1}
        \sum\nolimits_{k=1}^{K-1} A_{\text{accum}}(k)log(A_{\text{accum}}(k)) \\
        & + (1-A_{\text{accum}}(k))log(1-A_{\text{accum}}(k)),
\end{split}
\end{gather}

As show in Eq. (15) of the paper, the overall loss can be written as 
\begin{gather}
\begin{split}
    \Loss{} =& \Loss{photo} + w_1\Loss{ptc} + w_2\Loss{bg} +  w_3\Loss{flow} + w_4\Loss{vdiff}\\
    & + w_5\Loss{tv}(\Vgrid{\sigma}{\text{Can}}) + w_6\Loss{tv}(\Vgrid{\text{flow}}{t}) + w_7\Loss{tv}(\mathbf{D}),
\end{split}
\end{gather}
where $w_1$, $w_2$, $w_3$, $w_4$, $w_5$, $w_6$ and $w_7$ are weights to balance each component in the final coarse loss. In experiments, we set $w_1=1e-1$, $w_2=1e-2$, $w_3=1e-5$, $w_4=0.$, $w_5=1e-6$, $w_6=1e-3$ and $w_7=1e-1$ in coarse stage for all datasets, and set $w_1=1e-2$, $w_2=1e-3$, $w_3=1e-5$, $w_4=1e-5$, $w_5=1e-6$, $w_6=1e-3$ and $w_7=1e-1$
in fine stage for all datasets.

For progressive training in fine stage, we first train with 10 images with closest time steps with canonical step, and progressively add image with the closest time step every 60 iterations. 

\subsection{Training Strategy and Settings}
We propose a coarse-to-fine training strategy.
For the coarse stage, we set the expected voxel number to $67^3$, the scale factor of the \InpaintNet is 1.5, and the iteration number is 20k. As the purpose of the coarse stage is learning a proxy geometry to calculate the bounding box for the fine stage, we do not use \InpaintNet during the coarse stage. For the fine stage, the expected voxel number is set to $80^3$ for D-NeRF dataset and $70^3$ for HyperNeRF dataset. The scale factor of the \InpaintNet is 2.0 and the iteration number is 100k. We use Adam optimizaer and set learning rate 1e-1 for voxel grids and 1e-3 for networks. The training process of a scene takes around 1 day on a GeForce RTX 3090 GPU.
Our final model size on average is 260M for D-NeRF dataset and 440M for HyperNeRF dataset. The rendering peed is 7s per image for D-NeRF dataset and 15s per image for HyperNeRF dataset.

\subsection{Lego Complete Dataset}
We build a new dataset, named \emph{Lego Complete Dataset}, that animates the object \emph{LEGO} with three different motion patterns. For each scene, the test set is split into three categories to evaluate three abilities: \emph{space interpolation}, \emph{time interpolation}, and \emph{canonical interpolation} abilities. For space interpolation ability, we test four random views for each training time step. Also, we interpolate three time steps between two near training time steps to test the model's generalization ability over unseen times. Finally, to evaluate the learned canonical radiance field, we test 50 random views in canonical space. This results in $200 + 197 + 50=447$ test images in the test set. 

\section{More Results}

\subsection{Results on NHR dataset}

\begin{figure}[!t] \centering
    \includegraphics[width=\columnwidth]{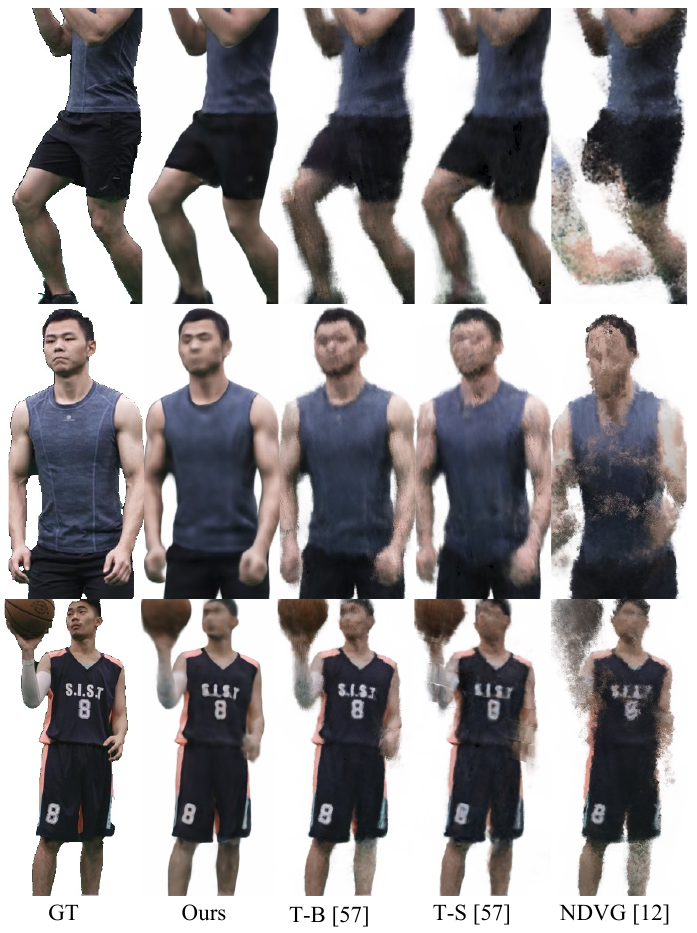}
    \vspace{-\baselineskip}
    \caption{\textbf{NHR dataset qualitative comparison.} We show some synthesized images on NHR dataset.
    } \label{fig:nhrcompare}
    \vspace{-2\baselineskip}
\end{figure}

We also test our method on NHR dataset \cite{wu2020_NHR}. We test all four scenes with 100 frames selecting 90\% views for train and 10\% views for test. Quantitative results are shown in \Tref{table:quantnhrdetail}, which shows ours achieving best results consistently. We show qualitative comparison in \Fref{fig:nhrcompare}, and our method could render clean and detailed images. We also visualize the learnt trajectories in \Fref{fig:nhrvistraj}. Use Acrobat to view animations.

\begin{figure}
    \centering
    \vspace*{-0.5\baselineskip}
    \captionsetup{type=figure}
    \subfloat{\animategraphics[trim={70 -30 170 1},every=3,width=0.54\linewidth,autoplay,loop]{4}{figures/render_traj_fine_last_sp1/traj_}{000}{099}}
    \hspace{0.01mm}
    \subfloat{\animategraphics[trim={250 -30 100 40},every=3,width=0.44\linewidth,autoplay,loop]{4}{figures/render_traj_fine_last_sp3/traj_}{000}{099}}

    \captionof{figure}{\footnotesize{\textbf{Trajectory visualization \textcolor{red}{Use Acrobat to view animations.}}}
    \label{fig:nhrvistraj}}
    \vspace*{-1.8\baselineskip}
\end{figure}

\begin{table*}[t!]
\setlength{\tabcolsep}{4pt} %
\centering
\caption{\small{
\textbf{Quantitative comparison.} Comparison of our method with others on LPIPS (lower is better) and PSNR/SSIM (higher is better) on eight dynamic scenes of the D-NeRF dataset.
}}
\label{table:quantsupp}
\resizebox{\textwidth}{!}{%
\begin{tabular}{lccccccccccccc}
\toprule
& & \multicolumn{3}{c}{Hell Warrior} & \multicolumn{3}{c}{Mutant} & \multicolumn{3}{c}{Hook} & \multicolumn{3}{c}{Bouncing Balls} \\
Methods & type
& PSNR$\uparrow$  & SSIM$\uparrow$ & LPIPS$\downarrow$
& PSNR$\uparrow$  & SSIM$\uparrow$ & LPIPS$\downarrow$
& PSNR$\uparrow$  & SSIM$\uparrow$ & LPIPS$\downarrow$
& PSNR$\uparrow$  & SSIM$\uparrow$ & LPIPS$\downarrow$\\
\cmidrule{1-1} \cmidrule(lr){2-2} \cmidrule(lr){3-5} \cmidrule(lr){6-8} \cmidrule(lr){9-11} \cmidrule(lr){12-14}
    TNeRF\,\cite{pumarola2021_dnerf_cvpr21} & N
    & 23.19 & 0.93 & 0.08
    & 30.56 & 0.96 & 0.04 
    & 27.21 & 0.94 & 0.06
    & 32.01 & 0.97 & 0.04 \\
    \midrule
    TiNeuVox-S ($100^3$)\,\cite{fang2022_TANV_arxiv} & NPC
    & 27.00 & 0.95 & 0.09
    & 31.09 & 0.96 & 0.05
    & 29.30 & 0.95 & 0.07
    & 39.05 & 0.99 & 0.06 \\
    TiNeuVox-B ($160^3$)\,\cite{fang2022_TANV_arxiv} & NPC
    & 28.17 & 0.97 & 0.07
    & 33.61 & 0.98 & 0.03
    & 31.45 & 0.97 & 0.05
    & 40.73 & 0.99 & 0.04 \\
    \midrule
    DNeRF\,\cite{pumarola2021_dnerf_cvpr21} & PC
    & 25.03 & 0.95 & 0.07
    & 31.29 & 0.98 & 0.03 
    & 29.26 & 0.97 & 0.12
    & 38.93 & \textbf{0.99} & 0.10 \\
    NDVG\,\cite{guo2022_NDVG_arxiv} & PC
    & 25.53 & 0.95 & 0.07
    & \textbf{35.53} & \textbf{0.99} & \textbf{0.01}
    & 29.80 & 0.97 & \textbf{0.04}
    & 34.58 & 0.97 & 0.11 \\
    Ours\,  & PC
    & \textbf{27.71} & \textbf{0.97} & \textbf{0.05}
    & 34.97 & 0.98 & 0.03
    & \textbf{32.29} & \textbf{0.98} & \textbf{0.04}
    & \textbf{40.02} & \textbf{0.99} & \textbf{0.04} \\
    \midrule
    Ours\_notv\,$^1$  & PC
    & 27.96 & 0.97 & 0.05
    & 35.26 & 0.98 & 0.03
    & 29.57 & 0.96 & 0.06
    & 40.40 & 0.99 & 0.05 \\
    Ours\_noup\,$^2$  & PC
    & 27.60 & 0.96 & 0.06
    & 34.15 & 0.98 & 0.04
    & 31.51 & 0.97 & 0.04
    & 38.89 & 0.99 & 0.05 \\
    Ours\_noinp\,$^3$  & PC
    & 27.15 & 0.96 & 0.06
    & 33.98 & 0.98 & 0.03
    & 31.77 & 0.97 & 0.04
    & 38.22 & 0.99 & 0.04 \\

\midrule
\midrule
& & \multicolumn{3}{c}{Lego} & \multicolumn{3}{c}{T-Rex} & \multicolumn{3}{c}{Stand Up} & \multicolumn{3}{c}{Jumping Jacks}  \\ 
Methods & type
& PSNR$\uparrow$  & SSIM$\uparrow$ & LPIPS$\downarrow$
& PSNR$\uparrow$  & SSIM$\uparrow$ & LPIPS$\downarrow$
& PSNR$\uparrow$  & SSIM$\uparrow$ & LPIPS$\downarrow$
& PSNR$\uparrow$  & SSIM$\uparrow$ & LPIPS$\downarrow$\\  
\cmidrule{1-1} \cmidrule(lr){2-2} \cmidrule(lr){3-5} \cmidrule(lr){6-8} \cmidrule(lr){9-11} \cmidrule(lr){12-14}
    TNeRF\,\cite{pumarola2021_dnerf_cvpr21} & N
    & 23.82 & 0.90 & 0.15
    & 30.19 & 0.96 & 0.13 
    & 31.24 & 0.97 & 0.02
    & 32.01 & 0.97 & 0.03 \\
    \midrule
    TiNeuVox-S ($100^3$)\, \cite{fang2022_TANV_arxiv} & NPC
    & 24.35 & 0.88 & 0.13
    & 29.95 & 0.96 & 0.06
    & 32.89 & 0.98 & 0.03
    & 32.33 & 0.97 & 0.04 \\
    TiNeuVox-B ($160^3$)\, \cite{fang2022_TANV_arxiv} & NPC
    & 25.02 & 0.92 & 0.07
    & 32.70 & 0.98 & 0.03
    & 35.43 & 0.99 & 0.02
    & 34.23 & 0.98 & 0.03 \\
    \midrule
    DNeRF\,\cite{pumarola2021_dnerf_cvpr21} & PC
    & 21.64 & 0.84 & 0.17
    & \textbf{31.76} & \textbf{0.98} & \textbf{0.04}
    & 32.80 & 0.98 & \textbf{0.02}
    & 32.80 & \textbf{0.98} & 0.04 \\
    NDVG\,\cite{guo2022_NDVG_arxiv}  & PC
    & 25.23 & 0.93 & \textbf{0.05}
    & 30.15 & 0.97 & 0.05
    & 34.05 & 0.98 & \textbf{0.02}
    & 29.45 & 0.96 & 0.08 \\
    Ours\,  & PC
    & \textbf{25.27} & \textbf{0.94} & \textbf{0.05}
    & 30.71 & 0.96 & \textbf{0.04}
    & \textbf{36.91} & \textbf{0.99} & \textbf{0.02}
    & \textbf{33.55} & \textbf{0.98} & \textbf{0.03} \\
    \midrule
    Ours\_notv\,$^1$  & PC
    & 24.33 & 0.89 & 0.11
    & 35.02 & 0.99 & 0.02
    & 37.01 & 0.99 & 0.02
    & 35.14 & 0.98 & 0.03 \\
    Ours\_noup\,$^2$  & PC
    & 25.20 & 0.93 & 0.06
    & 30.24 & 0.97 & 0.05
    & 35.47 & 0.98 & 0.02
    & 33.14 & 0.98 & 0.04 \\
    Ours\_noinp\,$^3$  & PC
    & 25.42 & 0.93 & 0.06
    & 30.00 & 0.97 & 0.04
    & 36.46 & 0.99 & 0.02
    & 31.24 & 0.98 & 0.04 \\

\bottomrule %

\multicolumn{10}{l}{$^1$~\footnotesize{not use the three total variation losses} \quad $^2$~\footnotesize{not up-sample the voxel grid} \quad $^3$~\footnotesize{not use the inpaint network}}

\end{tabular}
}
\vspace{-0.5\baselineskip}

\end{table*}

\subsection{Quantitative Results}

\paragraph{Detailed results for D-NeRF Dataset} We show more detailed results in \Tref{table:quantsupp} for D-NeRF dataset. As show in \Tref{table:quantsupp}, the inpaint network plays a relative important role. Also, the up-sample layer could improves the performance, which proves this layer learns to recover the details of the 3D voxel grid. This point gives some insight for image super-resolution direction, which is working in 3D dimension may benefit the 2D image tasks. Finally, the total variation losses helps the training to be more stable and get cleaner images.

\begin{figure}[!t] \centering
    \includegraphics[width=\columnwidth]{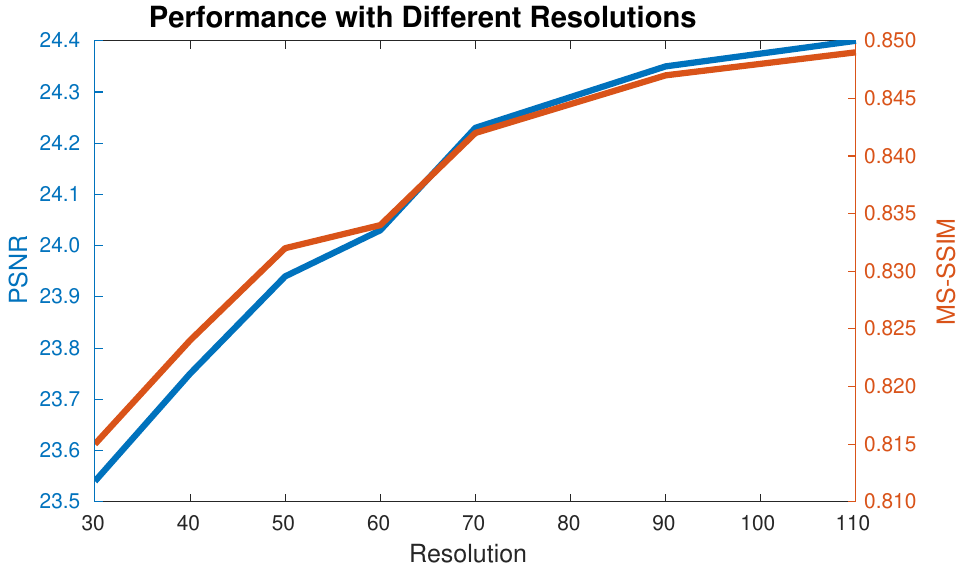}
    \vspace{-\baselineskip}
    \caption{\textbf{Performance with different resolutions}
    } \label{fig:resolution}
    \vspace{-2\baselineskip}
\end{figure}

\begin{table*}[!t]
\centering
\caption{
\textbf{More results.} Mean of Hell Warrior, Mutant, Hook and Bouncing Balls in DNeRF dataset.
}
\vspace{-0em}
\setlength\tabcolsep{3pt}
\label{tab:rebuttal}
    \resizebox{0.6\textwidth}{!}{
    \large
    \begin{tabular}{*{9}{c}}
        \toprule
       Method & ours &  w/o    & w/o      &       w/o       &    w/o   &  can-time &   w/o     & w/o $\Vgrid{\text{R}_{\text{Up}}}{}$\\
              &      &   $\Loss{tv}$   & $\Loss{flow}$ & $\Loss{vdiff}$  &  coarse  &  t at mid  & view dir  &      photo \\
        \midrule
        PSNR  & 33.75 & 33.30 & 33.74 & 33.57 & 31.63 & 34.09 & 22.81 & 31.56 \\
        SSIM  & 0.979 & 0.974 & 0.978 & 0.977 & 0.967 & 0.980 & 0.925 & 0.966 \\
        LPIPS & 0.040 & 0.048 & 0.039 & 0.041 & 0.058 & 0.037 & 0.096 & 0.057 \\
        \bottomrule
    \end{tabular}
    }
    \vspace{-0em}
\end{table*}

\paragraph{Resolutions}
We report performance with different resolutions on HyperNeRF dataset in \Fref{fig:resolution}. According to \Fref{fig:resolution}, the resolution of voxel grid plays an important role when it is relative small and the improvement of the performance decrease when resolution increasing.

\paragraph{Regularization terms} 
We study the effect of all regularization terms we proposed, including $\Loss{flow}$, $\Loss{vdiff}$ and $\Loss{tv}$ in \Tref{tab:rebuttal}. We could observe that all these three regularization terms has positive effect on the performance, but the improvement is minor. This proves the improvement of our method compared with others come from the forward warping desgin we proposed in the paper. Also, backward flow based method NDVG \cite{guo2022_NDVG_arxiv} use similar regularization terms with ours, and our method has clear advantage compared with NDVG \cite{guo2022_NDVG_arxiv}.

\paragraph{Training strategy} 
We also set our method without coarse stage training, show in \Tref{tab:rebuttal} (w/o coarse). The performance drops significantly that is reasonable, because the voxel grid covers bigger space without filtering with proxy geometry trained by coarse training. 

\paragraph{Canonical setup} 
In our method, we set canonical time to be the first frame for D-NeRF dataset to compare with \emph{physical canonical based method} and the middle frame for HyperNeRF dataset for better performance. Setting canonical time to be the middle frame helps improving the performance as the state of the middle frame geometry is closer to other time steps compared with the first frame. To prove this, we set canoical time to be middel frame in \Tref{tab:rebuttal} (can-time t at mid), and the PSNR improves slightly.

\paragraph{Ray direction modeling} 
According to Nerfies\cite{park2021_nerfies_iccv21},  we need to transfer the current view orientation to the directions in the canonical workspace. But we think using them directly is an acceptable solution for most papers in this area. We leave this as an open question for further study. We test with setting all directions to (0,0,1), the PSNR drops obviously in \Tref{tab:rebuttal} (w/o view dir). This proves the current solution works well to some extend.

\paragraph{Photmetric loss for $\Vgrid{\text{R}_{\text{Up}}}{}$} 
We use photometric terms on $\Vgrid{\text{R}_{\text{Up}}}{}$ to make sure the warped grid before inpainting could already render reasonable images. This make sure the UNet is actually doing `inpainting'. \Fref{fig:depthcompare} (bottom right) shows an example of how inpainting works. Also, we do observe inpainting and upsampling could refine grids. For videos and Fig. 8 in the main paper, the trajectories are reasonable which means we do learn trajectories without inpainting overfitting. We also test w/o photometric terms of $\Vgrid{\text{R}_{\text{Up}}}{}$ in \Tref{tab:rebuttal} (w/o $\Vgrid{\text{R}_{\text{Up}}}{}$ photo) and the performance drops as there is no direct supervise signal for trajectory training.

\subsection{Qualitative Results}

\begin{figure*}[!t] \centering
    \includegraphics[width=\textwidth]{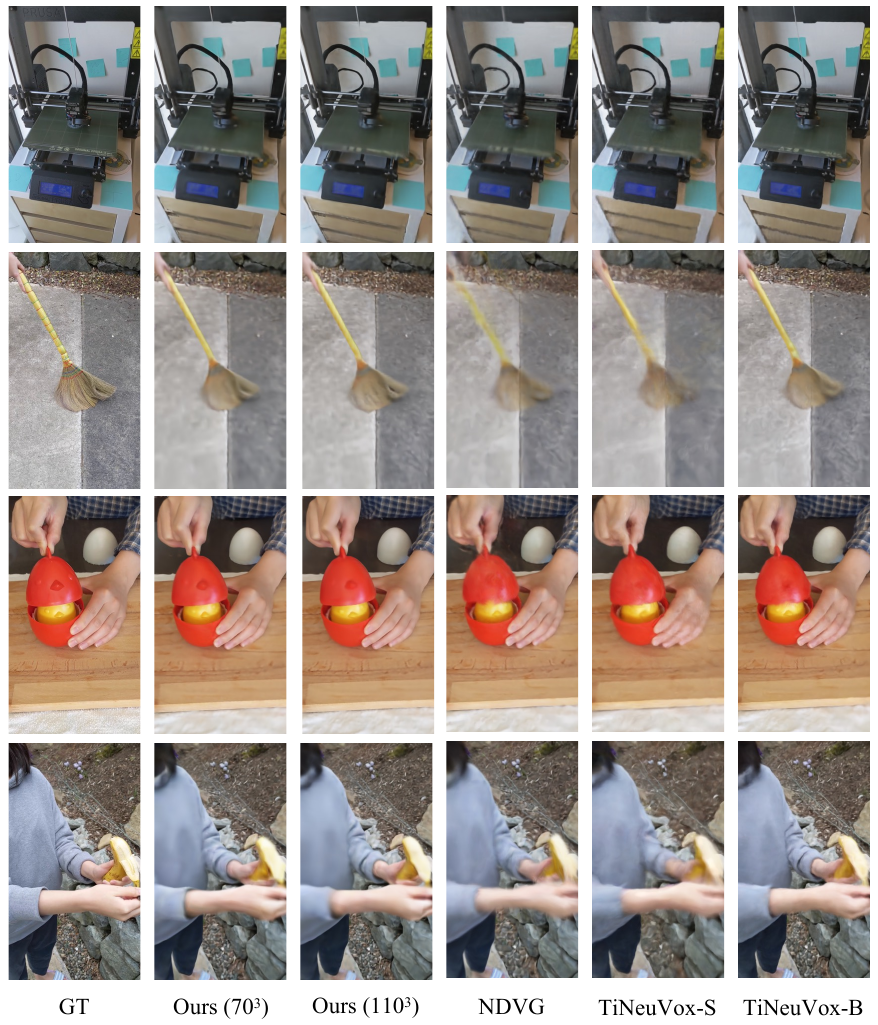}
    \vspace{-\baselineskip}
    \caption{\textbf{HyperNeRF dataset qualitative comparison.} We show some synthesized images on HyperNeRF dataset of our method with different resolutions and other methods.
    } \label{fig:hypercompare}
    \vspace{-2\baselineskip}
\end{figure*}

\begin{figure*}[!t] \centering
    \includegraphics[width=\textwidth]{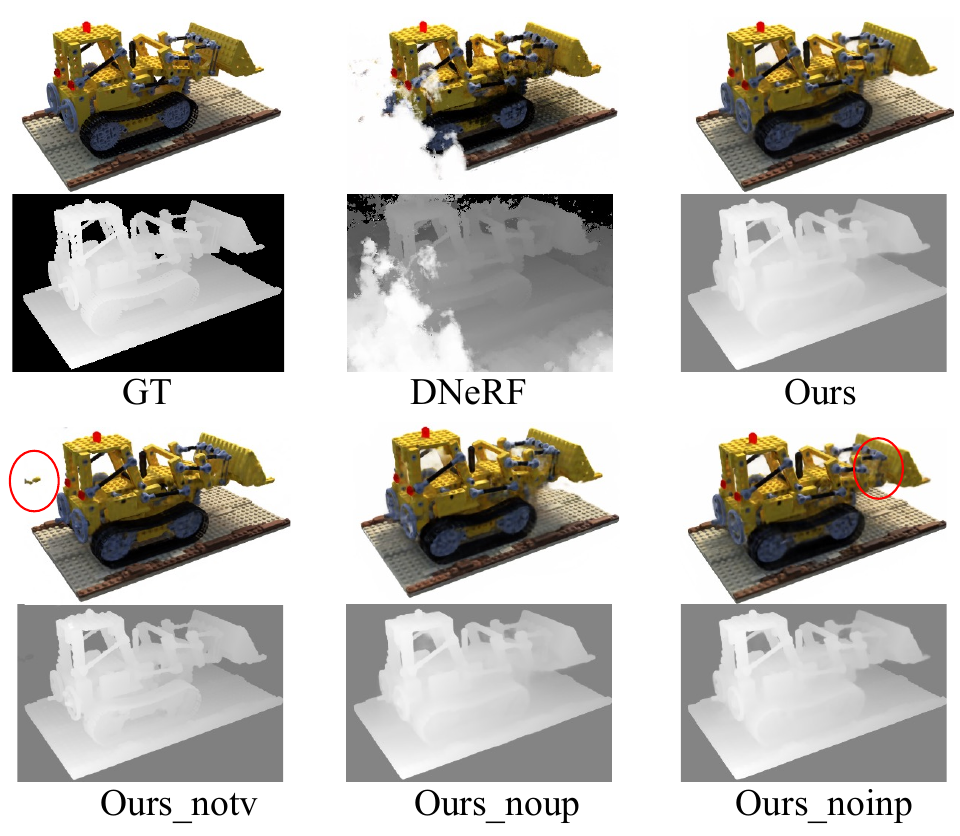}
    \vspace{-\baselineskip}
    \caption{\textbf{Lego Complete dataset qualitative comparison.} We show some synthesized images and depth rendered at the same test view selected from one scene of lego complete dataset with different settings.
    } \label{fig:depthcompare}
    \vspace{-2\baselineskip}
\end{figure*}

\begin{figure*}[!t] \centering
    \includegraphics[width=\textwidth]{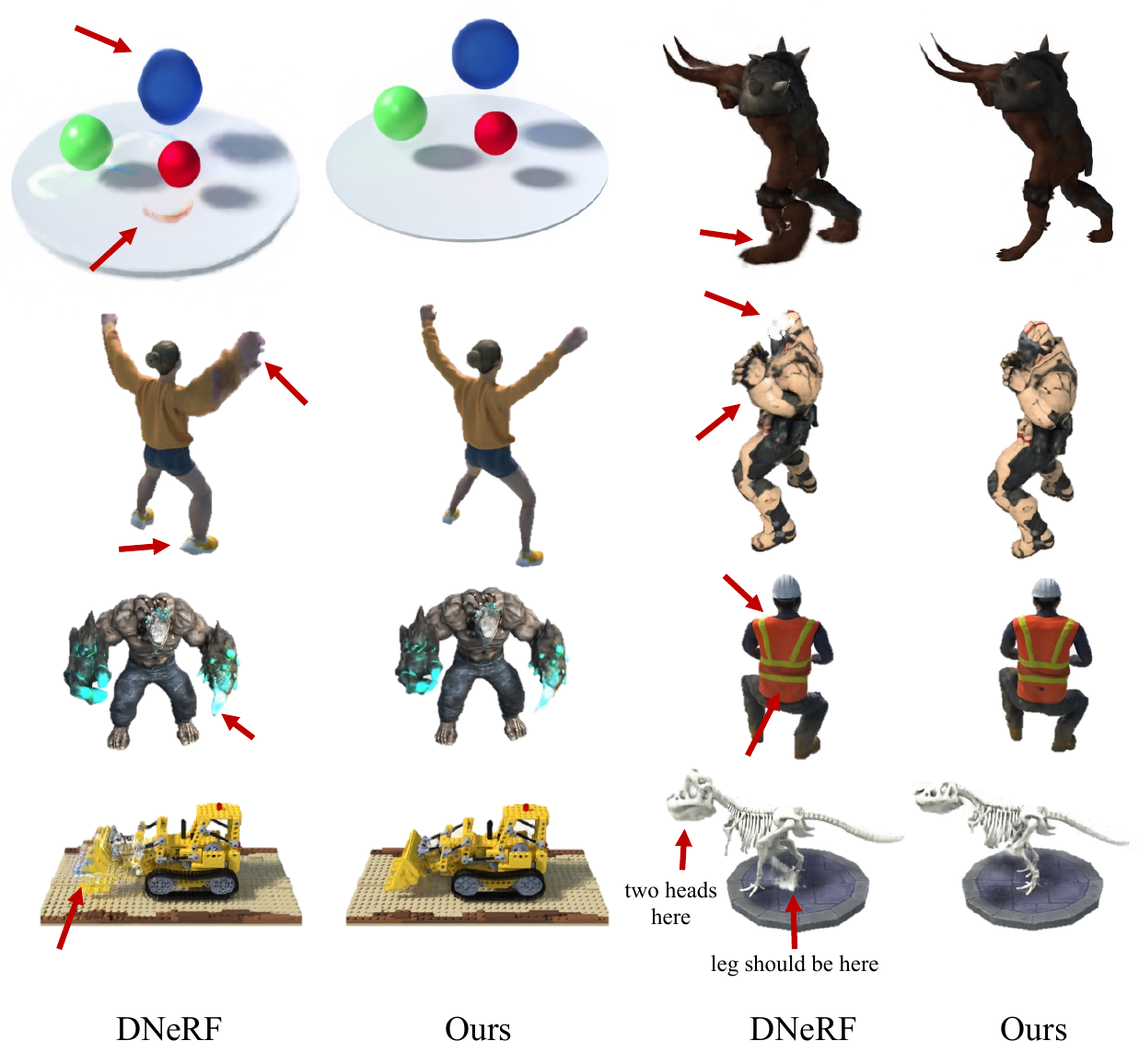}
    \vspace{-\baselineskip}
    \caption{\textbf{Canonical qualitative comparison.} We show canonical comparison of the DNeRF\cite{pumarola2021_dnerf_cvpr21} dataset. Note the differences highlighted by the arrows.
    } \label{fig:cancomparesupp}
    \vspace{-2\baselineskip}
\end{figure*}

We show rendered images of our method with different resolutions on HyperNeRF dataset in \Fref{fig:hypercompare}. With bigger resolutions, our method could recover more details, like the pattern of the 3D printer (first row), details of broom (second row) and details of the head in peel-banana (last row). Also, we provide more visual comparison with other methods in \Fref{fig:hypercompare}.

We show the rendered images and depths of our methods and D-NeRF\cite{pumarola2021_dnerf_cvpr21} in \Fref{fig:depthcompare}, compared with ground truth. Since we aims to synthesis dynamic scenes from monocular camera, which is a nontrivial problem, the model is highly possible to over-fit the training images. In \Fref{fig:depthcompare}, DNeRF is an example, which produce some clouds in the space which cause artifacts in other views. This is one of the reasons to build lego complete dataset, testing the abilities of the model to interpolate the time and space (including canonical space). Without total variation losses, we could get sharper depth but there may some noise points on the images. Without up-sample layer, the image is blurer (better zoom in for details). Without inpaint network, the rubber band of the lego arms disappears. The rubber band at this time step is stretched and this motion is non-rigid. This non-rigid motion would cause one-to-many issue, compared with rigid motions of mechanical structures of this lego. This proves our inpaint network could handle the one-to-many issue of forward warping.

We show the comparison of canonical image between D-NeRF\cite{pumarola2021_dnerf_cvpr21} and proposed method in \Fref{fig:cancomparesupp}. \Fref{fig:cancomparesupp} shows our method could recover correct canonical geometry in the canonical compared with DNeRF\cite{pumarola2021_dnerf_cvpr21}, which shows the power and potential of the forward warping.

We show more results in our video, including canonical comparison, trajectory visualization and other images render at novel views with different setting. Please refer to the supplement video for more information.

{\small
\bibliographystyle{ieee_fullname}
\bibliography{9_References}
}

\end{document}